\documentclass{article}

\usepackage[utf8]{inputenc} 
\usepackage[T1]{fontenc}    
\usepackage{hyperref}       
\usepackage{url}            
\usepackage{booktabs}       
\usepackage{amsfonts}       
\usepackage{nicefrac}       
\usepackage{microtype}      
\usepackage{xcolor}         
\usepackage{graphicx}
\usepackage{amsmath}
\usepackage{placeins}
\usepackage[ruled]{algorithm2e}
\usepackage[margin=1in]{geometry}
\usepackage{array}
\usepackage{algorithmic}
\usepackage{tabularx}
\def \n {\newline}

\title{ArtWhisperer: A Dataset for Characterizing Human-AI Interactions in Artistic Creations}

\author{%
  \hspace{-1em}Kailas Vodrahalli\thanks{Stanford University. Email: \texttt{kailasv@stanford.edu}}\hspace{-1em}
  \and James Zou\thanks{Stanford University. Email: \texttt{jamesz@stanford.edu}}\hspace{-1em}
}
\date{}

\begin{document}

\maketitle

\begin{abstract}
In this work, we investigate how people use text-to-image models to generate desired target images. To study this interaction, we created ArtWhisperer, an online game where users are given a target image and are tasked with iteratively finding a prompt that creates a similar-looking image as the target. 
Through this game, we recorded over 50,000 human-AI interactions; each interaction corresponds to one text prompt created by a user and the corresponding generated image. 
The majority of these are repeated interactions where a user iterates to find the best prompt for their target image, making this a unique sequential dataset for studying human-AI collaborations. 
In an initial analysis of this dataset, we identify several characteristics of prompt interactions and user strategies. 
People submit diverse prompts and are able to discover a variety of text descriptions that generate similar images. Interestingly, prompt diversity does not decrease as users find better prompts.
We further propose a new metric to quantify AI model \emph{steerability} using our dataset. We define steerability as the expected number of interactions required to adequately complete a task. We estimate this value by fitting a Markov chain for each target task and calculating the expected time to reach an adequate score.
We  quantify and compare AI steerability across different types of target images and two different models, finding that images of cities and nature are more steerable than artistic and fantasy images.
We also evaluate popular vision-language models to assess their image understanding and ability to incorporate feedback.
These findings provide insights into human-AI interaction behavior, present a concrete method of assessing AI steerability, and demonstrate the general utility of the ArtWhisperer dataset.
\end{abstract}
\section{Introduction}\label{sec:introduction}
Direct human interaction with AI models has become widespread following a number of technical innovations improving the quality of text-to-text \cite{brown2020language,ouyang2022training,anil2023palm} and text-to-image models \cite{Rombach_2022_CVPR,ramesh2022hierarchical}, enabling the public release of high-quality AI-based services like ChatGPT \cite{chatGPT}, Bard \cite{Bard}, and Midjourney \cite{Midjourney}. 
These models have seen rapid interest and adoption largely due to the ability of the general public to interact with and steer the AI in diverse contexts including engineering, creative writing, art, education, medicine, and law \cite{dakhel2023github,nguyen2022empirical,ippolito2022creative,cetinic2022understanding,qadir2023engineering,cascella2023evaluating,reuters_article_2023}. 

A key challenge in developing these models is aligning their output to human inputs. This is made challenging by the broad domain of use cases as well as the diverse prompting styles of different users. Many approaches can be categorized as ``prompt engineering,'' where specific strategies for prompting are used to steer a model \cite{oppenlaender2022prompt,liu2022design,zhou2022large,wei2022chain,white2023prompt}. Great success has also been found by fine-tuning models using relatively small datasets to follow human instructions \cite{ouyang2022training}, respond in a specific style \cite{hu2021lora}, or behave differently to specified prompts \cite{zhou2022learning,gal2022image}.

\begin{figure*}
    \centering
    \includegraphics[width=5.5in]{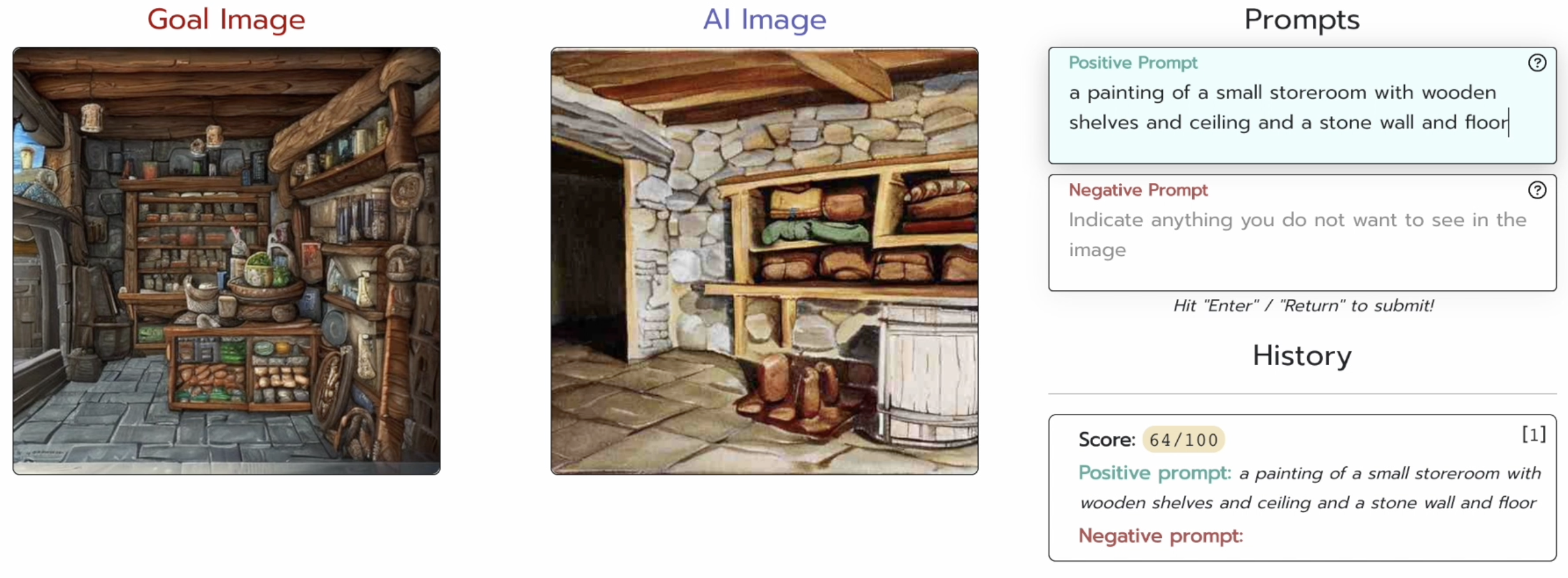}
    \caption{Interface of the ArtWhisperer game. Prompts entered on right. Target (goal) image and player-generated image on left. Previous prompts and scores are displayed in the lower right.}
    \label{fig:game_interface}
\end{figure*}

In this work, we take interest in the fact that human interaction with these models is often an iterative process.
We develop a dataset to study this interaction. The dataset is collected through an interactive game we created where players try to find an optimal prompt for a given task (see Figure~\ref{fig:game_interface}). In particular, we focus on text-to-image models and ask the player to generate a similar image (\textit{AI Image}) to a given \textit{target image}. The player is allowed to iterate on their prompt, using the previously generated image(s) as feedback to help them adjust their prompt. A score is also provided as feedback to help the user calibrate how ``close'' they are to a similar image.

Using this setup, we collected data on 51,026 interactions from 2,250 players across 191 unique target images. The target images were selected from a diverse set of AI-generated and natural images. 
We also collected a separate dataset of 4,572  interactions, 140 users, and 51 unique target images in a more controlled setting 
to assess the robustness of our findings. 

Based on this data, we find several interesting patterns in how people interact with AI models. 
Players discover a diverse set of prompts that all result in images similar to the target. To discover these prompts, players typically make small, iterative updates to their prompts. Each update improves their image with a moderate success rate ($40-60\%$ for most target images).
Based on these findings, we define and evaluate a metric for model steerability using the stopping time of an empirical Markov model. We use this metric to assess steerability across image categories and across two AI models. 

\textbf{Our contributions} 
We release a public dataset on human interactions with an AI model. To our knowledge, this is the first such dataset showing repeated interactions of people with a text-to-image model to accomplish specified tasks. We also provide an initial analysis of this data and propose a simple-to-calculate metric for assessing model steerability. Additionally, we use our dataset and this steerability metric to evaluate the ability of vision-language models to utilize feedback.
Our dataset and associated code is made available at \href{https://github.com/kailas-v/ArtWhisperer}{https://github.com/kailas-v/ArtWhisperer}.

\textbf{Related Works}\label{sec:related_works}
Human-AI interaction datasets for text-to-text and text-to-image models typically focus on single interactions and generally do not provide users with a specific task. 
Public text-to-image interaction datasets typically contain the generated AI images and/or prompts \cite{huggingface_dataset, wang2022diffusiondb} and optionally some form of human preference rating \cite{pressmancrowson2022,wu2023alsd,xu2023imagereward,kirstain2023pick}. These datasets generally rely on scraping online repositories like Lexica \cite{lexica} or Discord servers focused on AI art. Though some of these datasets include metadata that may allow for reconstruction of prompt iteration, there is no guarantee the user has the same desired output in mind over the iteration.
Public text-to-text interaction datasets are much more limited as the best performing models are generally accessible only through APIs with no public user interaction datasets. While some researchers have investigated how human-AI interaction for text-to-text can be improved through various tools \cite{wu2022promptchainer,wu2022ai}, the amount of data collected is limited and not publicly available. There are also repositories containing prompt strategies for various tasks \cite{bach2022promptsource} but no human interaction component.

We seek to rectify two of the shortcomings of the existing datasets--namely, that they do not contain extended interactions as the user attempts to steer the AI, and they do not have a predefined goal. In our work, we maintain a controlled environment for human users where we allow extended interactions towards a fixed goal. 
As shown by our initial analysis, our dataset may enable deeper understanding of user prompting strategies, assessing model steerability, and evaluating vision-language AI models.
\section{Interaction Game}\label{sec:game_description}
\begin{figure*}
    \centering
    \includegraphics[width=5.5in]{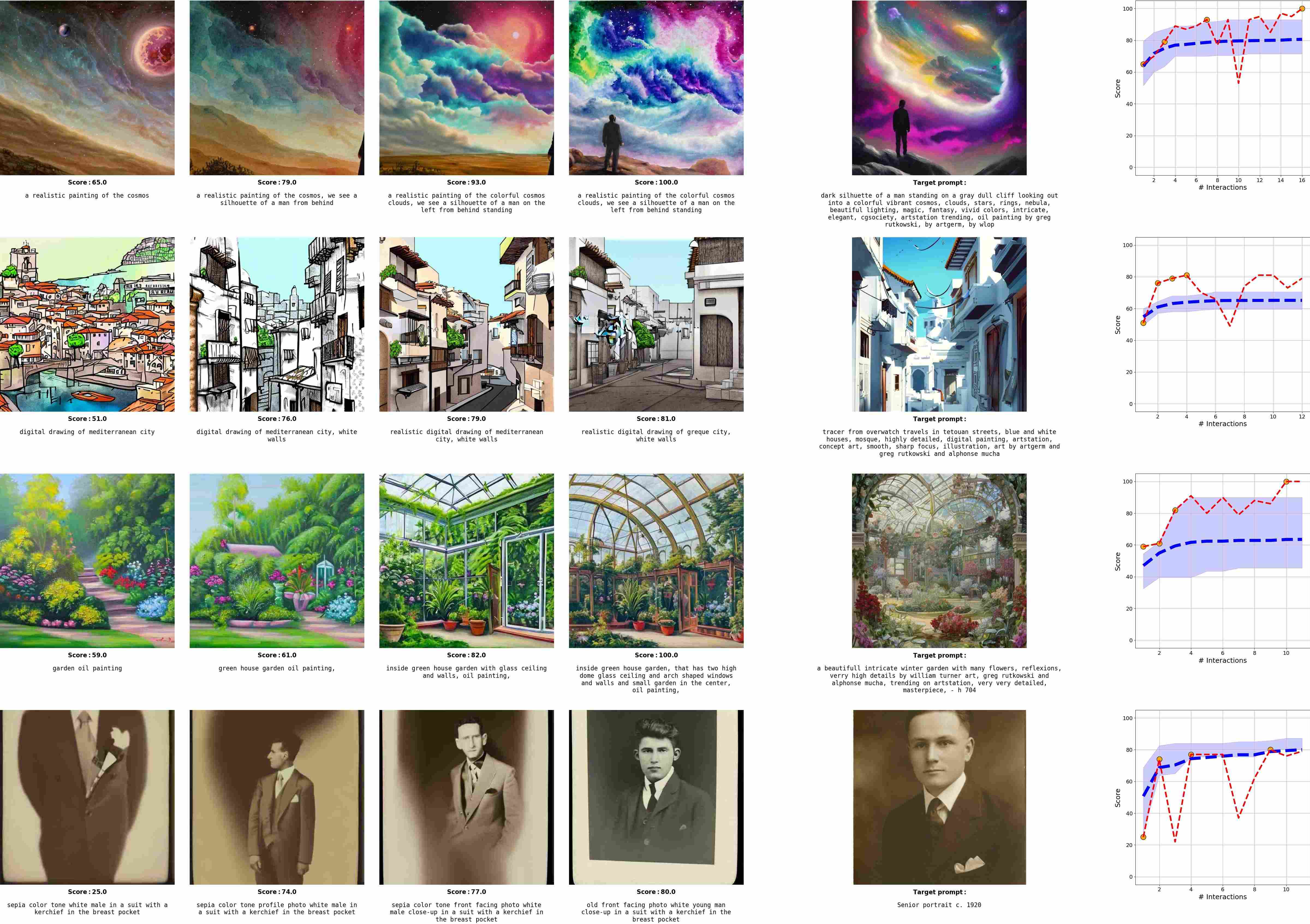}
    \caption{Example user trajectories. In each row, we show (1) a given user's prompts, (2) the target image (rightmost image), and (3) a plot of this target image's average score trajectory across users (blue), this  user's full score trajectory (red), and the displayed images (orange).
    }
    \label{fig:ex_target_images}
\end{figure*}

In our game, players are shown a target image (see Figures~\ref{fig:ex_target_images} and~\ref{fig:diverse_prompt_examples}). 
Players are also given a limited interface to a text-to-image model, Stable Diffusion (SD) v2.1 model \cite{rombach2022high}.
In particular, players can enter a ``positive prompt'' (describes the desired content) and a ``negative prompt'' (describes what should be omitted) to steer the AI model. All models hyperparameters including the seed are fixed. 
Upon inputting a prompt, the player is shown the image generated by the AI model, along with a similarity score between their generated image and the target image. The interface is shown in Figure~\ref{fig:game_interface}.

\subsection{How Target Images are Selected}\label{sec:target_image_selection}
We randomly sample target images from two sources. The first is a collection of Wikipedia pages, and the second is a dataset of prompts AI artists have used with SD \cite{huggingface_dataset}. 
In addition to sampling target images, we need to ensure the task is feasible to users. We do not allow users to adjust the seed or other parameters of the model, so we need to ensure the selected model parameters can generate reasonably similar images to the target image. We find that selecting an appropriate random seed is sufficient, and fix all other model parameters (see Appendix~\ref{app:sec:target_image_examples} for examples of generated images and Appendix~\ref{sec:game_technical_details} for details and discussion).

\vspace{-0.5em}
\paragraph{Wikipedia Images}
A collection of 35 Wikipedia pages on various topics including art, nature, cities, and various people. A full list of pages sampled from is provided in Appendix~\ref{app:tbl:wikipedia_images}. From these pages, we scraped 670 figures licensed under the Creative Commons license. These figures were then filtered by which had captions, as well as which images were JPG or PNG images (i.e., not animated, and not PDF files), resulting in 557 images. 

For each of the 557 images, we first resize and crop the image to size $512 \times 512$. The Wikipedia caption is used as the ground truth ``prompt''. Let the image-caption pair be denoted as $(t_i, p_i^*)$. We sample the model on 50 random seeds, with $p_i^*$ as the prompt input. This generates a set of 50 images: $S_i = \{(x_i, s_i): i=1,\dots,50\}$ for generated image $x_i$ and seed $s_i$. Let $C(x)$ denote the CLIP image embedding \cite{radford2021learning} of an image $x$. Then we select the seed as $s_{i^*}$, where 
\begin{align}
    i^* := \min_{i=1,\dots,50} \left|\left| \frac{C(x_i)}{||C(x_i)||_2}-\frac{C(t_i)}{||C(t_i)||_2} \right|\right|_2 \label{eqn:wiki_seed}
\end{align}
    
Here, $s_{i^*}$ is selected to minimize the distance to the target image given the target prompt.

\paragraph{AI-Generated Images}
A collection of 2,000 AI-art prompts are randomly sampled from the Stable Diffusion Prompts dataset \cite{huggingface_dataset}. For each prompt, $p_i^*$, we generate two sets of images. As before, we use 50 unique random seeds to select the seed, $s_{i^*}$ and an additional 10 random seeds to use for selecting the generated target image (so in total, we use 60 unique random seeds): the first set, $S_{i,1} = \{(x_{i,1}, s_i): i=1,\dots,10\}$ and $S_{i,2} = \{(x_{i,2}, s_i): i=1,\dots,50\}$. We select the target image, $t_{i_1^*}$, from $S_{i,1}$:

\begin{align}
    i_1^* := \min_{i=1,\dots,10} \text{median} & \left( \left\{ \left|\left| \frac{C(x_{i,1})}{||C(x_{i,1})||_2}-\frac{C(x_{j,2})}{||C(x_{j,2})||} \right|\right|_2 :  j=1,\dots,50\right\} \right) \label{eqn:ai_target_image}
\end{align}

We select the random seed, $s_{i_2^*}$, using $t_{i_1^*}$ and $S_{i,2}$, with
\begin{align}
    i_2^* := \min_{i=1,\dots,50} \left|\left| \frac{C(x_{i,2})}{||C(x_{i,2})||_2}-\frac{C(t_{i_1^*})}{||C(t_{i_1^*})||_2} \right|\right|_2 \label{eqn:ai_seed}
\end{align}

Here, $t_{i_1^*}$ is chosen to be more representative of the types of images we may expect given the fixed prompt, $p_i^*$. This is because $t_{i_1^*}$ is selected to be close to the center of the sampled images, $S_{i,2}$. The intuition for selecting $s_{i_2^*}$ is the same as selecting $s_{i^*}$ for the Wikipedia images.

\subsection{Scoring Function}\label{sec:scoring_function}
To provide feedback to players, we created a scoring function to assess the similarity of a player's generated image and the target image. We define the scoring function as 
\begin{align}
    score(x, t) = \max(0, \min(100, alpha \cdot \left|\left| \frac{C(x)}{||C(x)||_2} - \frac{C(t)}{||C(t)||_2} \right|\right|_2 + \beta)) \label{eqn:score_fn}
\end{align}
for generated image $x$, target image $t$, and constants $\alpha, \beta$. Note the range of $score(x, t)$ is integers in the interval $[0, 100]$. Details on how $\alpha, \beta$ are selected parameters are provided in Appendix~\ref{app:scoring_function}.

While this scoring function is often reasonable, it does not always align with the opinions of a human user. To assess how well $score(x, t)$ follows a user's preferences, we acquire ratings from a subset of users (see \textit{ArtWhisperer-Validation} in Section~\ref{sec:data_overview}). We find $\text{score}(x, t)$ has a Pearson correlation coefficient of 0.579 indicating reasonable agreement. Further assessment is performed in Section~\ref{sec:automated_scoring} and discussed at length in Appendix~\ref{app:human_rating_discussion}.

\subsection{Dataset Overview}\label{sec:data_overview}
We collected two datasets: \textit{ArtWhisperer} and \textit{ArtWhisperer-Validation}. We use \textit{ArtWhisperer} for most analysis and results; for some of the results in Sections~\ref{sec:scoring_function}, \ref{sec:analysis},and \ref{sec:automated_scoring}, we also use \textit{ArtWhisperer-Validation} (when referenced). 
Data was collected from March-May 2023. IRB approval was obtained.

\begin{figure}
    \centering
    \includegraphics[width=3in]{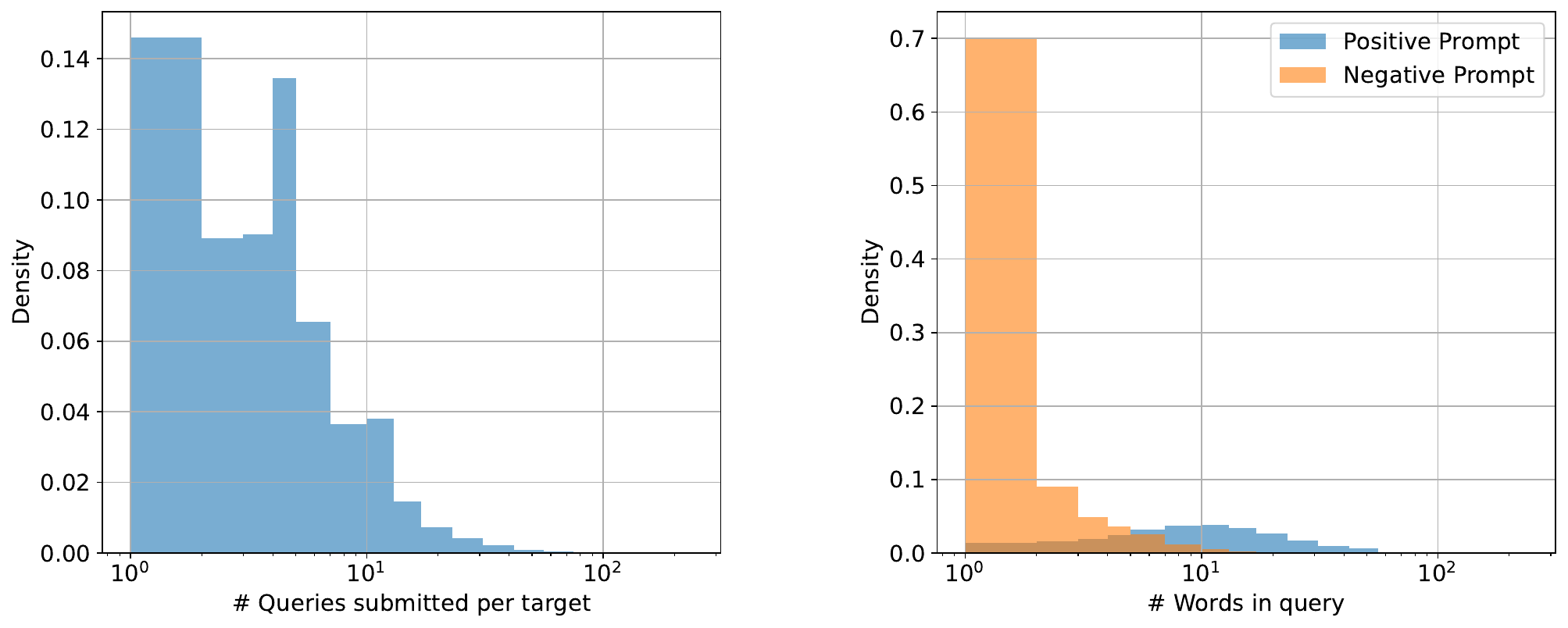}
    \caption{Left: Distribution of \# of user queries per target image (average queries per image is 9.18). Right: Distribution of the \# of words submitted in a query (average words submitted is 20.02 and 2.32 for positive and negative prompts respectively).
    }
    \label{fig:query_distribution}
\end{figure}

\textbf{\textit{ArtWhisperer}:} 
A public version of our game was released online, with three new target images released daily.
We collected data from consenting users playing the game. Users were not paid.
Users were anonymous and we only collected data related to the prompts submitted to ensure privacy of users. While we expect some users played the game across multiple days, we did not track them. A summary of the \textit{ArtWhisperer} dataset is provided in Table~\ref{tbl:overview_of_data}. In total, we have 2,250 (potentially non-unique) players corresponding to 51,026 interactions across 191 target images. 
Players interacted with the model SD v2.1.
In Figure~\ref{fig:query_distribution}, we plot the number of queries submitted by players across different target images. 

\textbf{\textit{ArtWhisperer-Validation}:} 
The game (with a near identical interface) was also released as a controlled user study to paid crowd workers on Prolific \cite{prolific}. The crowd workers were compensated at a rate of $\$12.00$ per hour for roughly 20 minutes of their time. Workers played the game across 5 randomly selected target images from a pre-selected subset of 51 target images chosen to have diverse content. Workers were also asked to rate each of their images on a scale of 1-10 (i.e., self-scoring their generated images). In total, we collected data on 4,572 interactions, corresponding to 140 users and 51 unique target images across two different diffusion models, SD v2.1 and SD v1.5.  
Additional details and demographic information are provided in Appendix~\ref{app:validation_data_overview}. 

\begin{table*}[ht]
\caption{\textit{ArtWhisperer} Dataset Overview. Each row contains summary data for a different subset of the dataset. Subsets may overlap. Similar information for \textit{ArtWhisperer-Validation} is in Appendix~\ref{app:validation_data_overview}.
} 
\label{tbl:overview_of_data}
\vskip -0.3in
\begin{center}
\begin{tabular}{>{\centering\arraybackslash}m{1.2cm}|>{\centering\arraybackslash}m{1.1cm}|>{\centering\arraybackslash}m{1.1cm}|>{\centering\arraybackslash}m{1.3cm}|>{\centering\arraybackslash}m{1.1cm}|>{\centering\arraybackslash}m{1.2cm}|>{\centering\arraybackslash}m{4cm}}
\toprule
\small \# Players & \small \# Target Images & \small \# Interactions & \small Average \n\# Prompts & \small Average Score & \small Median Duration & \small Category \\ 
\hline\hline
\small \textbf{2250} & \small \textbf{191} & \small \textbf{51026} & \small \textbf{9.29} & \small \textbf{58.93} & \small \textbf{18 s} & \small \textbf{Total} \\
\hline\hline
\small 377 & \small 25 & \small 3884 & \small 8.65 & \small 56.70 & \small 19 s & \small Contains famous person? \\
\hline
\small 353 & \small 32 & \small 3785 & \small 8.26 & \small 61.64 & \small 21 s & \small Contains famous landmark? \\
\hline
\small 2005 & \small 140 & \small 40290 & \small 9.24 & \small 59.83 & \small 18 s & \small Contains man-made content? \\
\hline
\small 1177 & \small 58 & \small 18255 & \small 10.93 & \small 57.21 & \small 17 s & \small Contains people? \\
\hline
\small 344 & \small 77 & \small 6972 & \small 8.81 & \small 62.01 & \small 20 s & \small Is real image? \\
\hline
\small 2140 & \small 103 & \small 43524 & \small 9.42 & \small 58.37 & \small 17 s & \small Is AI image? \\
\hline
\small 1483 & \small 82 & \small 24913 & \small 9.14 & \small 59.45 & \small 17 s & \small Is art? \\
\hline
\small 623 & \small 29 & \small 7297 & \small 9.14 & \small 53.77 & \small 18 s & \small Contains nature? \\
\hline
\small 160 & \small 14 & \small 1355 & \small 7.28 & \small 65.74 & \small 19 s & \small Contains city? \\
\hline
\small 1239 & \small 39 & \small 15872 & \small 9.91 & \small 56.74 & \small 16 s & \small Is fantasy? \\
\hline
\small 618 & \small 19 & \small 8359 & \small 10.51 & \small 57.88 & \small 17 s & \small Is sci-fi or space? \\

\bottomrule
\end{tabular} 
\end{center}
\vskip -0.1in
\end{table*}

\section{Prompt Diversity}\label{sec:prompt_diversity}
We quantify prompt diversity by looking at the distribution of prompts in the text embedding space. In particular, we use the CLIP text embedding \cite{radford2021learning}, though we do find the choice of embedding is not particularly important for our results (see Appendix~\ref{app:alternative_text_embeddings}).

\subsection{Diverse prompts used for high scores}
\begin{figure*}
    \centering
    \includegraphics[width=5.5in]{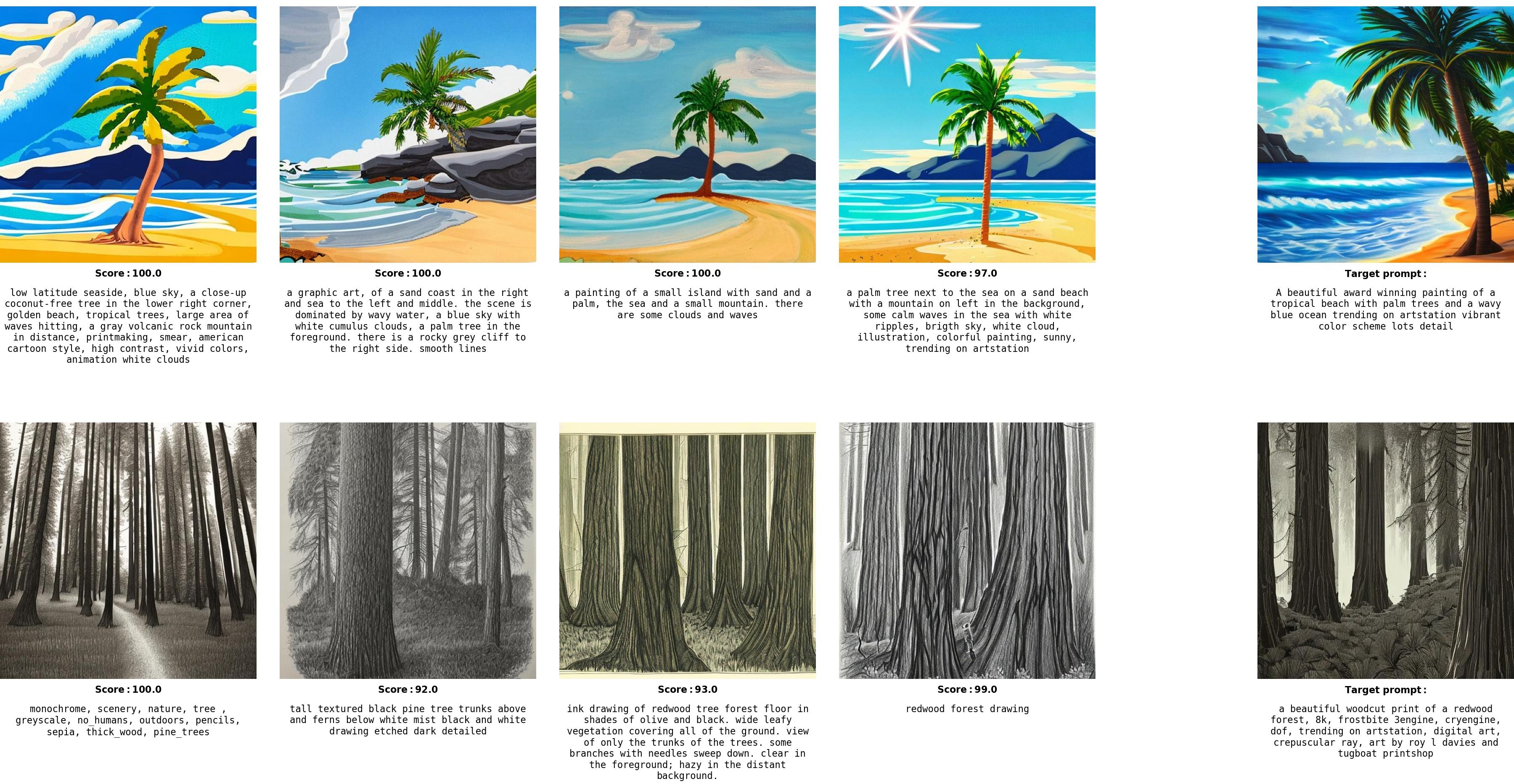}
    \caption{Diverse, high-scoring prompt submissions from different users. Target image in rightmost column.
    }
    \label{fig:diverse_prompt_examples}
\end{figure*}

First, we find that users do not converge in their prompt design but rather achieve similarly high scores through a diverse set of prompts. This result confirms the need for models to align to a wide range of user writing styles and descriptive techniques, and suggests a tradeoff between alignment to multiple users and specificity of a prompt (e.g., to support more users, individual prompts will necessarily have less descriptive ability).
Examples of such diverse prompts are shown in Figure~\ref{fig:diverse_prompt_examples}.

\begin{figure}
    \centering
    \includegraphics[width=3in]{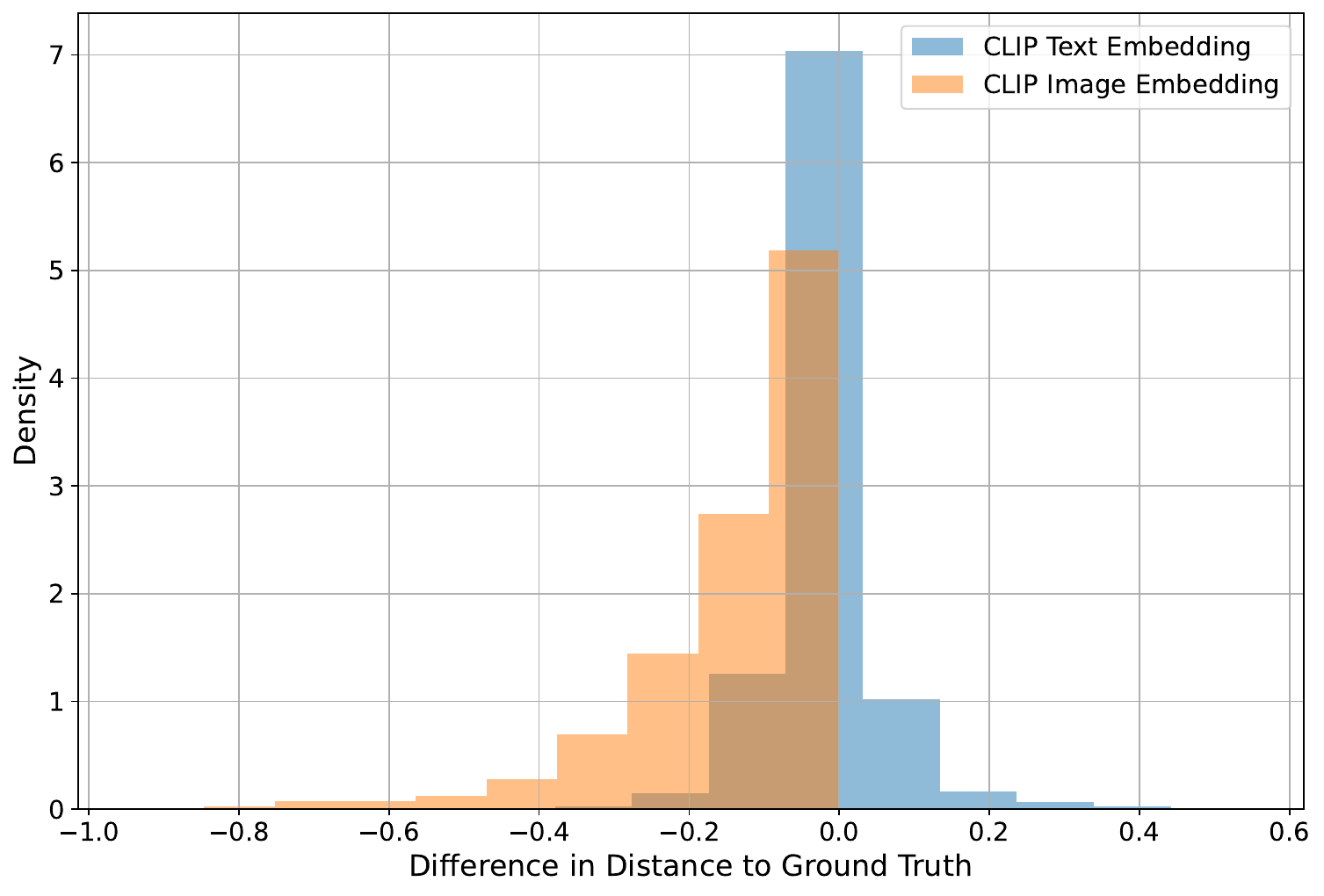}
    \caption{Difference of distance from the first prompt to ground truth and distance from the last (best) prompt to ground truth for CLIP text (blue) and CLIP image embeddings (orange).
    }
    \label{fig:text_and_image_embedding_variation}
\end{figure}

We quantify this finding in Figure~\ref{fig:text_and_image_embedding_variation}, where we plot two metrics defined as follows. Let $z_0, z_n$ be normalized embeddings of the initial and best prompt/image found by a user. Let $z^*$ be the normalized embedding of the target prompt/image. We define the difference in embedding distance to ground truth as $||z_n-z^*||_2 - ||z_0 - z^*||_2$. In blue, we use the CLIP text embeddings of the prompts; in orange, we use the CLIP image embeddings of the generated images. We note two findings here: (1) the metric applied to the image embeddings is guaranteed to be non-positive as the embedding distance is monotonically decreasing with the score, and (2) the metric applied to the text embeddings is apparently symmetric around $0$, indicating that unlike the image embedding, distance in the text embedding space does not monotonically decrease with score. Together, these findings illustrate that users tend to discover diverse prompts and \emph{do not converge in their prompt design}. 

Additionally, we find the distribution of prompts does not converge. In the left of Figure~\ref{fig:prompt_diversity}, we plot the distribution of distances between the first prompt (in blue) and the last prompt (in orange) to the average prompt for the corresponding target image. Despite the average score improving from 51.9 to 70.3 (out of 100) indicating a significant improvement in score, prompt diversity does not significantly diminish. That is, users do not converge to similar prompts to achieve high scores. Similar analysis of the image embedding space suggests image diversity \emph{decreases} (Figure~\ref{fig:text_and_image_embedding_variation}). 

\subsection{People submit similar prompts throughout their interaction}\label{sec:prompt_strategies}
Second, we find that people do not explore a wide range of prompt designs even when their initial prompts are not performing well. This suggests the lack of convergence in prompt design is inherent to the users’ preference to describe an image.
It also suggests that user initialization (i.e., the first submitted prompt) is critical, and that online personalization may be possible (to adapt to the user's writing style) given the more stable nature of an individual's prompt design for a given target image.

In the center of Figure~\ref{fig:prompt_diversity}, we plot the distribution of the standard deviation of prompts for users (blue) and for permuted users (orange). Permuted users are generated by sampling from all prompts for a given target image uniformly, using the same distribution of number of prompts as for real users. The gap between the two distributions shows that individuals do not randomly sample prompts each interaction, but base new prompts off of previously submitted prompts ($\text{p-value}<10^{-10}$, t-test for independent variables). An analysis of how scores change between adjacent prompts shows that this strategy has a moderate success rate and improves the score $40-60\%$ of the time, with an average rate of $48.6\%$ (note that score changes $<1$ are counted as unchanged; this occurs $10.2\%$ of the time).

\subsection{People have similar prompt styles across images}
Finally, we find that people have a measurable writing style or prompt design that appears across images; however, this prompt design is predominantly informed by the target image rather than any consistent style. This implies that online personalization \emph{across different target images} may require a large number of images to be effective (across a small set of individual target images, the prompt design and writing style may still change significantly). This is in contrast to our earlier finding that suggests personalization for a given image may be possible given the relative stability of a user's prompt design for a given target image.

\begin{figure*}
    \centering
    \includegraphics[width=5.5in]{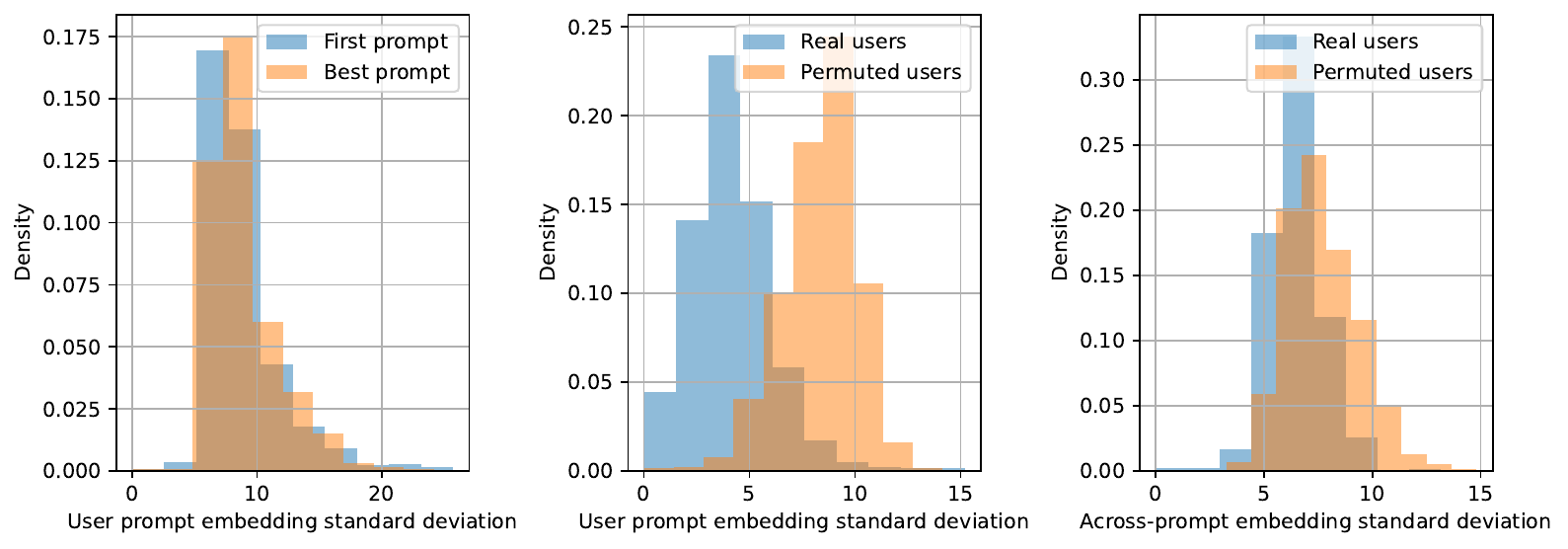}
    \caption{Left: Users submit diverse (across users) prompts, both at beginning and end of interaction. 
    Center: Individual users do not submit diverse prompts.
    Right: Users have different prompting styles.
    }
    \label{fig:prompt_diversity}
\end{figure*}

We quantify user style by computing the difference (in the CLIP text embedding space) between the average prompt of a given user and the average prompt across all users for a given target image. To quantify style variation for a user, we then compute the standard deviation of the user style across the target images the user generated.
In the right of Figure~\ref{fig:prompt_diversity}, we plot the distribution of user style variation for real users (blue) and permuted users (orange). Permuted users are generated by randomly sampling user styles. This allows us to test whether users have a consistent prompting style. 
We find users do indeed have specific styles of prompting ($\text{p-value}<10^{-10}$, t-test for independent variables). However, the difference is  seemingly not large, suggesting that while user style may a component to prompting, other factors related to the target image may be more important.

\section{Model Steerability}\label{sec:steerability}
Model steerability refers to the ability of a user to steer a model towards a desired outcome. Measuring steerability has great utility, as it enables quantitative tracking of the human usability of generative models. However, there is no current consensus on how to measure AI steerability. A common approach is to simply measure performance of a model on standardized dataset evaluations \cite{jahanian2019steerability, openai2023gpt4}. While this can enable comparisons between tasks and models, this approach does not allow for the feedback loop present when humans interact with a model. Steerability can also be measured qualitatively based on user assessment of their experience interacting with the AI \cite{chung2022talebrush}.

Here, we propose a simple yet informative measure to assess model steerability.
This metric is general across model types and data modalities, and we are able to use it here to compare across image categories and models.
We then analyze this measure across different subgroups of images and across two different Stable Diffusion models: SDv2.1 and the older SDv1.5 \cite{Rombach_2022_CVPR}.

\subsection{Measuring steerability}

As discussed in Section~\ref{sec:prompt_strategies}, users typically engage with the model through clusters of similar prompts. They typically start with an initial base prompt and proceed to make multiple incremental modifications to it. 
We use this observation as a basis for creating a steerability metric. 
We define a Markov chain between scores. Each node is a score with edges connecting to the subsequent score. To make this tractable for empirical analysis, we bin scores into five groups: $[0,20], [21,40], [41,60], [61,80], [81,100]$. We use the expected time taken to reach the last score bin, $[81,100]$, as our steerability score (i.e., the stopping time to reach an adequate score).

For each target image, we calculate the empirical transition probability matrix between binned scores using all the players' data for that image. We then calculate the \textit{steerability score} for the given target image by running a Monte Carlo simulation to estimate stopping time, as defined above.
To assess steerability across a group of images, we average steerability score across all images in the group.

\subsection{Analysis}\label{sec:analysis}
\begin{figure}
    \centering
    \includegraphics[width=3in]{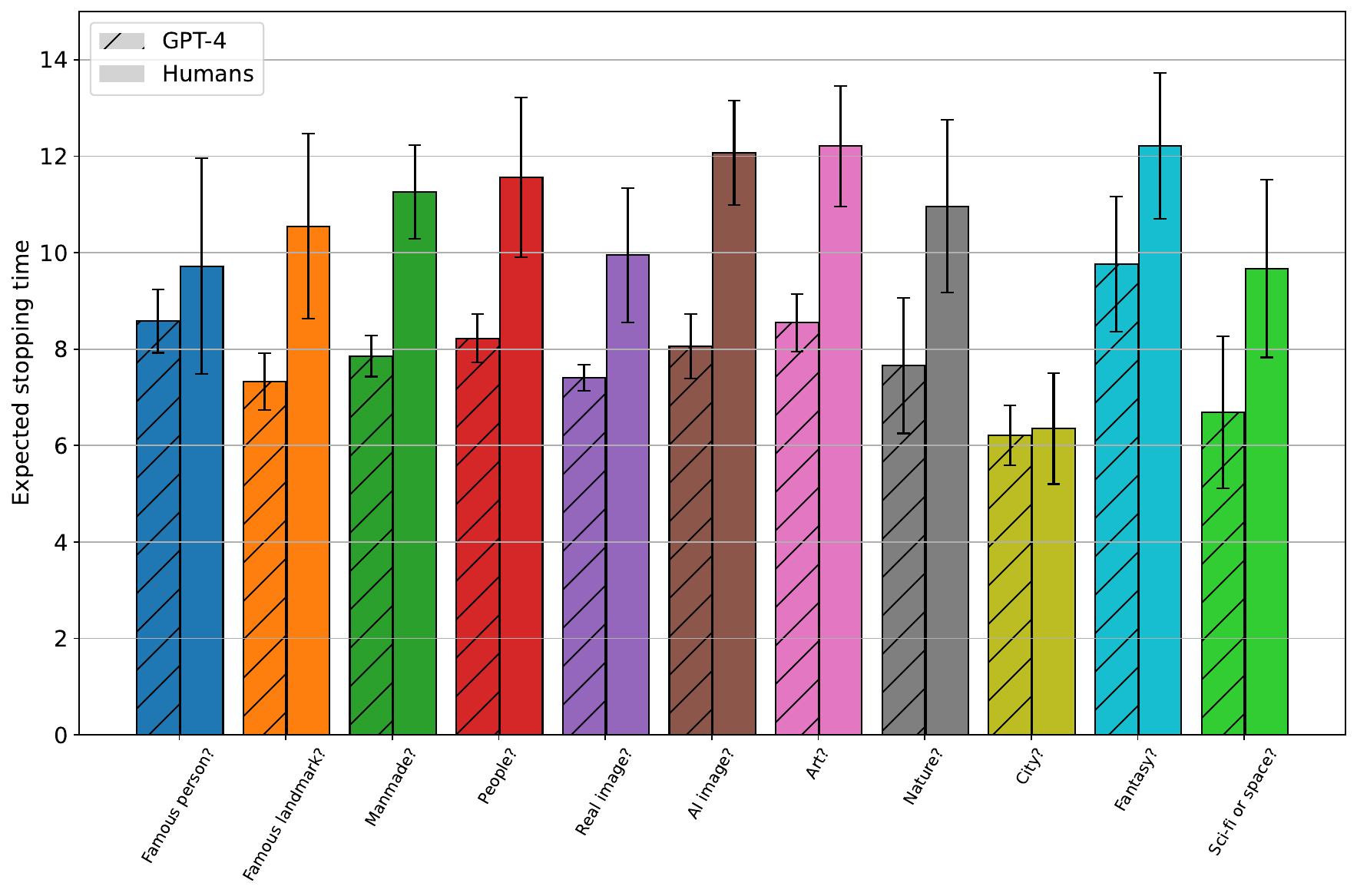}
    \vspace{-0.2in}
    \caption{Steerability across image groups for human users and GPT-4 (smaller indicates increased steerability). Bars show average expected stopping time across images in the image group; error bars show standard error. }
    \label{fig:steerability_across_images}
\end{figure}

In Figure~\ref{fig:steerability_across_images}, we plot the steerability score across image groups. Error bars show the standard error. For examples of steerability scores for individual images, see Appendix~\ref{app:individual_image_steerability}. 
We find that images containing famous people or landmarks, real images (not AI generated), contain cities, or contain nature are the most steerable. AI-generated images, fantasy images, and images of human art are the least steerable. There are a few possible explanations. The model we are assessing here, SDv2.1, as well as its text encoder OpenCLIP, are trained on subsets of LAION5B \cite{schuhmann2022laionb}. The contents of LAION5B are predominantly real world images, indicating why these images may be more steerable (i.e., text describing these types of images may have a better encoding). Moreover, the prompts for AI-generated images and fantasy images generally include specific internet artists and/or art styles which may not be known to most users making achieving the desired target image more difficult. Another potential reason is the distribution of images chosen for each category. Clearly, there are ``easier'' and ``more difficult'' images in each category; part of the reason for smaller stopping time may be the sample of images chosen rather than the actual image category.

Using the \textit{ArtWhisperer-Validation} data, we also compare steerability across two models: SDv2.1 and SDv1.5. Across most image categories, we observe a similar steerability. Images of nature, sci-fi or space, and real images have the largest differences in steerability between the two models; SDv2.1 is more steerable in all three cases. This suggests that SDv2.1 may be more steerable for natural images as well as sci-fi images, and is similarly steerable for other kinds of images including AI-generated artwork. One explanation may be that most of our users were not aware of certain prompting strategies that help models generate more aesthetic images or certain art styles; it is possible that for experienced users, AI art images may be more steerable, and differences between models may be magnified if, for example, a user is experienced working with one particular model.
More discussion is provided in Appendix~\ref{app:steerability_across_models}.

\subsection{Justification for automated score}\label{sec:automated_scoring}
One limitation of our steerability metric comes from the method of scoring user-submitted prompts.  
Ideally, we would assess steerability with a user's personal preferences. 
As mentioned in Section~\ref{sec:scoring_function}, the scores and human ratings have a positive correlation. 
Here, we use the human ratings from \textit{ArtWhisperer-Validation} instead of our score function to assess steerability. We compute the steerability score across both models and across image groups. Generally, the steerability scores change little. In all but two cases (SDv2.1 on sci-fi and space images; SDv1.5 on nature images), the human rating-based steerability score remains within a 95\% confidence interval of the score-based steerability score. While our score function may not perfectly capture human preferences, the steerability score we generate appears to be robust to these issues. Further discussion is included in Appendix~\ref{app:human_rating_discussion}.

\section{Vision-Language Model Evaluation}\label{sec:vlm_evaluation}
We also use the ArtWhisperer dataset to evaluate vision-language models (VLMs) for their ability to incorporate feedback. In particular, we have two VLMs, GPT-4 and Gemini \cite{openai2023gpt4, team2023gemini}, play the ArtWhisperer game across all target images. In this context, the models are interacting with a ``tool'' -- an SD model. However, we can also consider this SD model as a proxy for a user whose preferences  the model must adapt to over time.

A system prompt is crafted to inform the model about the game. A starting prompt is used to query the model for an initial prompt. The generated prompt is evaluated using the ArtWhisperer game pipeline -- an image is generated and then scored. In the ``Feedback'' mode, this generated image and the ArtWhisperer score is then fed back to the VLM with a request for an updated prompt. This process is repeated until the model attains a perfect score or 20 attempts have been made. In  ``No feedback'' mode, the VLM is not given any feedback and is just queried repeatedly without any conversation history given. More details on the prompts used are included in Appendix~\ref{app:vlm_feedback_evaluation}. 

In Figure~\ref{fig:vlm_trajectory}, we compare the average score (across target images) trajectory of the VLM models over time. Here, the score is the ArtWhisperer game score, where larger values indicate a better prompt. For each of GPT-4 and Gemini, we have two trajectories. ``GPT-4'' and ``Gemini'' represent the prompting methodology described above and further detailed in Appendix~\ref{app:vlm_feedback_evaluation}. The ``No feedback'' plots represent the average score across queries as well, as each query is independent; this is a baseline measure of the VLM's ability to prompt Stable Diffusion given an image with no feedback. 

When given feedback, GPT-4 consistently improves as it receives more feedback, with minimal improvement after about 15 rounds of feedback. This indicates the model is able to incorporate feedback well. In contrast, Gemini does not improve with feedback. This is expected -- the Gemini model we evaluated on was not trained for multi-turn conversations (an API for such a model was not released at the time of writing). 

In Figure~\ref{fig:steerability_across_images}, we compare steerability of the Stable Diffusion model with respect to GPT-4 and humans. Across all tasks, Stable Diffusion is at least as steerable with respect to GPT-4 as it is with humans. This result holds even when we normalize for initial score (i.e., to reduce the effect of baseline prompt writing ability and just examine ability to use feedback), indicating that GPT-4 is better at adapting to feedback compared to human users. This additional plot and more discussion is included in Appendix~\ref{app:vlm_feedback_evaluation}.

\begin{figure}
    \centering
    \includegraphics[width=3in]{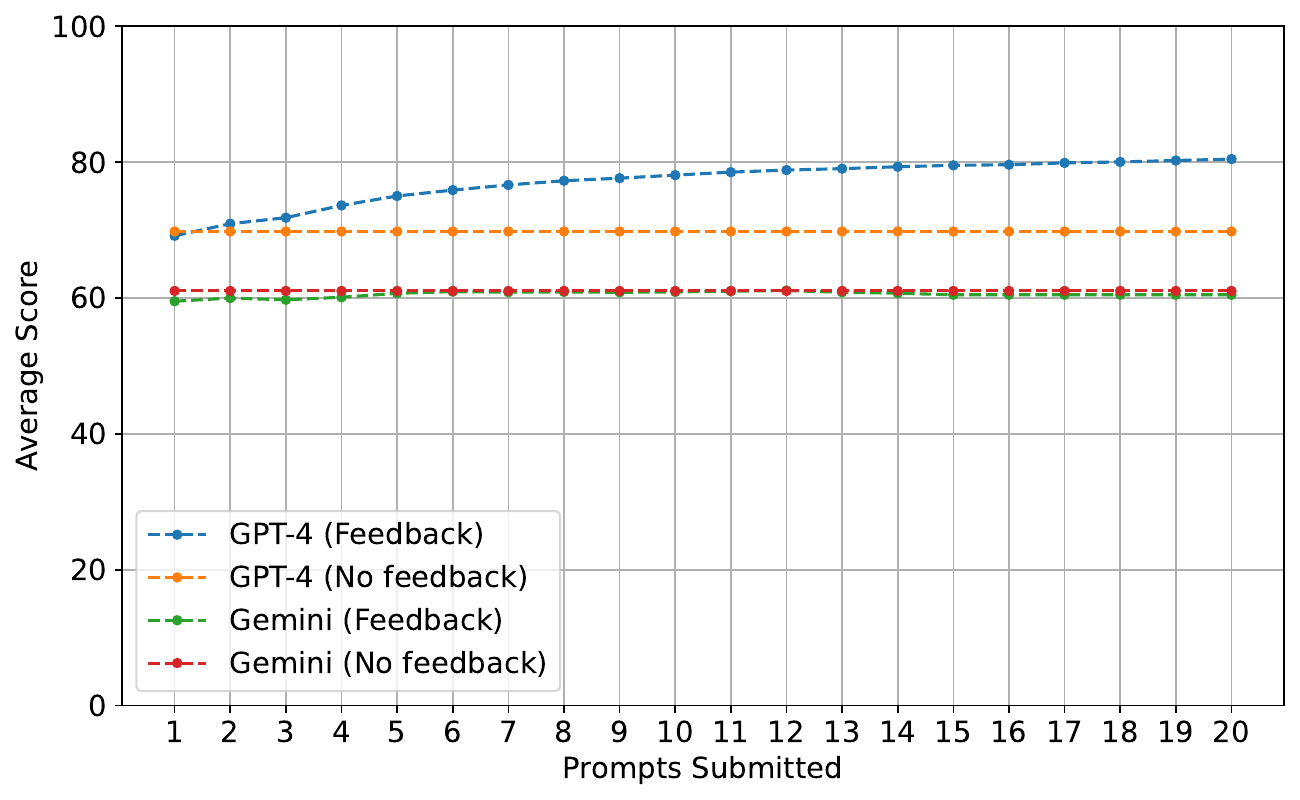}
    \caption{Averaged trajectory of GPT-4 and Gemini. ``No feedback'' does not use any feedback from previously generated images.}
    \label{fig:vlm_trajectory}
\end{figure}
\section{Discussion}\label{sec:discussion}
As demonstrated in our analysis, the \textit{ArtWhisperer} and \textit{ArtWhisperer-Validation} datasets can provide insights into  user prompting strategies and enables us to assess model steerability for individual tasks and groups of tasks. What makes our dataset particularly useful is the controlled interactive environment, where users work toward a fixed goal, that we capture data in. 

One of the most exciting use cases we see for our dataset is to create synthetic humans for prompt generation. For example, similar to the method described in Promptist \cite{hao2022optimizing}, we imagine fine-tuning a large language model with our dataset to generate prompt trajectories (i.e., rather than an optimized prompt) using similar exploration strategies as a human prompter. These synthetic prompters could be based on multimodal models like OpenFlamingo \cite{awadalla2023openflamingo} or text-only models and use score-feedback to condition the trajectory generation. 
As an initial proof-of-concept, we fine-tuned a MT0-large model \cite{muennighoff2022crosslingual} model on our dataset and found the fine-tuned model can indeed behave similarly to human users (see Appendix~\ref{app:synthetic_prompter}). It is also feasible that a VLM like the ones assessed in Section~\ref{sec:vlm_evaluation}
 could be used as synthetic humans through few-shot prompting, obviating the need for fine-tuning. These synthetic prompters have several uses: 

\begin{enumerate}
    \item Automating measurement of text-to-image model steerability by using synthetic users in place of real human prompters. While we believe our proposed steerability metric is effective, its main limitation currently is the requirement for human annotations. 
    \item Incorporating steerability in the objective function for text-to-image models. By representing steerability as a function of synthetic users, it becomes possible to explicitly optimize a model for steerability.
    \item Generating human readable image captions that are compatible with a Stable Diffusion model by using the synthetic prompter to optimize the token representation of the prompt. This is related to \cite{zhu2023collaborative}, where the authors use models to revise prompts.
\end{enumerate}

Additionally, our dataset can be used for further analysis on human prompting strategies beyond what we discussed in the paper. For example, one question we only touched upon is whether we can compare human prompters to automated prompt optimization methods (e.g., do humans behave similar to some gradient-based optimization approach in the prompt embedding space?). There are also potential uses for crafting better image similarity metrics using the human ratings we collected.

Finally, as discussed in Section~\ref{sec:vlm_evaluation}, our dataset and methodology is useful for assessing vision-language models. In particular, we are able to capture the ability of models to utilize feedback to adapt to a given user or tool.

\FloatBarrier

\bibliographystyle{plain}
\bibliography{references}

\begin{thebibliography}{10}

\bibitem{anil2023palm}
Rohan Anil, Andrew~M. Dai, Orhan Firat, Melvin Johnson, Dmitry Lepikhin,
  Alexandre Passos, Siamak Shakeri, Emanuel Taropa, Paige Bailey, Zhifeng Chen,
  Eric Chu, Jonathan~H. Clark, Laurent~El Shafey, Yanping Huang, Kathy
  Meier-Hellstern, Gaurav Mishra, Erica Moreira, Mark Omernick, Kevin Robinson,
  Sebastian Ruder, Yi~Tay, Kefan Xiao, Yuanzhong Xu, Yujing Zhang,
  Gustavo~Hernandez Abrego, Junwhan Ahn, Jacob Austin, Paul Barham, Jan Botha,
  James Bradbury, Siddhartha Brahma, Kevin Brooks, Michele Catasta, Yong Cheng,
  Colin Cherry, Christopher~A. Choquette-Choo, Aakanksha Chowdhery, Clément
  Crepy, Shachi Dave, Mostafa Dehghani, Sunipa Dev, Jacob Devlin, Mark Díaz,
  Nan Du, Ethan Dyer, Vlad Feinberg, Fangxiaoyu Feng, Vlad Fienber, Markus
  Freitag, Xavier Garcia, Sebastian Gehrmann, Lucas Gonzalez, Guy Gur-Ari,
  Steven Hand, Hadi Hashemi, Le~Hou, Joshua Howland, Andrea Hu, Jeffrey Hui,
  Jeremy Hurwitz, Michael Isard, Abe Ittycheriah, Matthew Jagielski, Wenhao
  Jia, Kathleen Kenealy, Maxim Krikun, Sneha Kudugunta, Chang Lan, Katherine
  Lee, Benjamin Lee, Eric Li, Music Li, Wei Li, YaGuang Li, Jian Li, Hyeontaek
  Lim, Hanzhao Lin, Zhongtao Liu, Frederick Liu, Marcello Maggioni, Aroma
  Mahendru, Joshua Maynez, Vedant Misra, Maysam Moussalem, Zachary Nado, John
  Nham, Eric Ni, Andrew Nystrom, Alicia Parrish, Marie Pellat, Martin Polacek,
  Alex Polozov, Reiner Pope, Siyuan Qiao, Emily Reif, Bryan Richter, Parker
  Riley, Alex~Castro Ros, Aurko Roy, Brennan Saeta, Rajkumar Samuel, Renee
  Shelby, Ambrose Slone, Daniel Smilkov, David~R. So, Daniel Sohn, Simon
  Tokumine, Dasha Valter, Vijay Vasudevan, Kiran Vodrahalli, Xuezhi Wang,
  Pidong Wang, Zirui Wang, Tao Wang, John Wieting, Yuhuai Wu, Kelvin Xu, Yunhan
  Xu, Linting Xue, Pengcheng Yin, Jiahui Yu, Qiao Zhang, Steven Zheng,
  Ce~Zheng, Weikang Zhou, Denny Zhou, Slav Petrov, and Yonghui Wu.
\newblock Palm 2 technical report, 2023.

\bibitem{awadalla2023openflamingo}
Anas Awadalla, Irena Gao, Josh Gardner, Jack Hessel, Yusuf Hanafy, Wanrong Zhu,
  Kalyani Marathe, Yonatan Bitton, Samir Gadre, Shiori Sagawa, Jenia Jitsev,
  Simon Kornblith, Pang~Wei Koh, Gabriel Ilharco, Mitchell Wortsman, and Ludwig
  Schmidt.
\newblock Openflamingo: An open-source framework for training large
  autoregressive vision-language models.
\newblock {\em arXiv preprint arXiv:2308.01390}, 2023.

\bibitem{bach2022promptsource}
Stephen~H Bach, Victor Sanh, Zheng-Xin Yong, Albert Webson, Colin Raffel,
  Nihal~V Nayak, Abheesht Sharma, Taewoon Kim, M~Saiful Bari, Thibault Fevry,
  et~al.
\newblock Promptsource: An integrated development environment and repository
  for natural language prompts.
\newblock {\em arXiv preprint arXiv:2202.01279}, 2022.

\bibitem{Bard}
\url{https://http://bard.google.com}, 2023.

\bibitem{brown2020language}
Tom Brown, Benjamin Mann, Nick Ryder, Melanie Subbiah, Jared~D Kaplan, Prafulla
  Dhariwal, Arvind Neelakantan, Pranav Shyam, Girish Sastry, Amanda Askell,
  et~al.
\newblock Language models are few-shot learners.
\newblock {\em Advances in neural information processing systems},
  33:1877--1901, 2020.

\bibitem{cascella2023evaluating}
Marco Cascella, Jonathan Montomoli, Valentina Bellini, and Elena Bignami.
\newblock Evaluating the feasibility of chatgpt in healthcare: an analysis of
  multiple clinical and research scenarios.
\newblock {\em Journal of Medical Systems}, 47(1):33, 2023.

\bibitem{cetinic2022understanding}
Eva Cetinic and James She.
\newblock Understanding and creating art with ai: review and outlook.
\newblock {\em ACM Transactions on Multimedia Computing, Communications, and
  Applications (TOMM)}, 18(2):1--22, 2022.

\bibitem{chatGPT}
\url{https://chat.openai.com}, 2023.

\bibitem{chung2022talebrush}
John Joon~Young Chung, Wooseok Kim, Kang~Min Yoo, Hwaran Lee, Eytan Adar, and
  Minsuk Chang.
\newblock Talebrush: sketching stories with generative pretrained language
  models.
\newblock In {\em Proceedings of the 2022 CHI Conference on Human Factors in
  Computing Systems}, pages 1--19, 2022.

\bibitem{dakhel2023github}
Arghavan~Moradi Dakhel, Vahid Majdinasab, Amin Nikanjam, Foutse Khomh, Michel~C
  Desmarais, and Zhen Ming~Jack Jiang.
\newblock Github copilot ai pair programmer: Asset or liability?
\newblock {\em Journal of Systems and Software}, 203:111734, 2023.

\bibitem{devlin2018bert}
Jacob Devlin, Ming-Wei Chang, Kenton Lee, and Kristina Toutanova.
\newblock Bert: Pre-training of deep bidirectional transformers for language
  understanding.
\newblock {\em arXiv preprint arXiv:1810.04805}, 2018.

\bibitem{gal2022image}
Rinon Gal, Yuval Alaluf, Yuval Atzmon, Or~Patashnik, Amit~H Bermano, Gal
  Chechik, and Daniel Cohen-Or.
\newblock An image is worth one word: Personalizing text-to-image generation
  using textual inversion.
\newblock {\em arXiv preprint arXiv:2208.01618}, 2022.

\bibitem{hao2022optimizing}
Yaru Hao, Zewen Chi, Li~Dong, and Furu Wei.
\newblock Optimizing prompts for text-to-image generation.
\newblock {\em arXiv preprint arXiv:2212.09611}, 2022.

\bibitem{he2016deep}
Kaiming He, Xiangyu Zhang, Shaoqing Ren, and Jian Sun.
\newblock Deep residual learning for image recognition.
\newblock In {\em Proceedings of the IEEE conference on computer vision and
  pattern recognition}, pages 770--778, 2016.

\bibitem{hu2021lora}
Edward~J Hu, Yelong Shen, Phillip Wallis, Zeyuan Allen-Zhu, Yuanzhi Li, Shean
  Wang, Lu~Wang, and Weizhu Chen.
\newblock Lora: Low-rank adaptation of large language models.
\newblock {\em arXiv preprint arXiv:2106.09685}, 2021.

\bibitem{ippolito2022creative}
Daphne Ippolito, Ann Yuan, Andy Coenen, and Sehmon Burnam.
\newblock Creative writing with an ai-powered writing assistant: Perspectives
  from professional writers.
\newblock {\em arXiv preprint arXiv:2211.05030}, 2022.

\bibitem{jahanian2019steerability}
Ali Jahanian, Lucy Chai, and Phillip Isola.
\newblock On the" steerability" of generative adversarial networks.
\newblock {\em arXiv preprint arXiv:1907.07171}, 2019.

\bibitem{khandelwal2022simple}
Apoorv Khandelwal, Luca Weihs, Roozbeh Mottaghi, and Aniruddha Kembhavi.
\newblock Simple but effective: Clip embeddings for embodied ai.
\newblock In {\em Proceedings of the IEEE/CVF Conference on Computer Vision and
  Pattern Recognition}, pages 14829--14838, 2022.

\bibitem{kirstain2023pick}
Yuval Kirstain, Adam Polyak, Uriel Singer, Shahbuland Matiana, Joe Penna, and
  Omer Levy.
\newblock Pick-a-pic: An open dataset of user preferences for text-to-image
  generation.
\newblock {\em arXiv preprint arXiv:2305.01569}, 2023.

\bibitem{lexica}
\url{https://lexica.art/}, 2023.

\bibitem{liu2022few}
Haokun Liu, Derek Tam, Mohammed Muqeeth, Jay Mohta, Tenghao Huang, Mohit
  Bansal, and Colin~A Raffel.
\newblock Few-shot parameter-efficient fine-tuning is better and cheaper than
  in-context learning.
\newblock {\em Advances in Neural Information Processing Systems},
  35:1950--1965, 2022.

\bibitem{liu2022design}
Vivian Liu and Lydia~B Chilton.
\newblock Design guidelines for prompt engineering text-to-image generative
  models.
\newblock In {\em Proceedings of the 2022 CHI Conference on Human Factors in
  Computing Systems}, pages 1--23, 2022.

\bibitem{loshchilov2017decoupled}
Ilya Loshchilov and Frank Hutter.
\newblock Decoupled weight decay regularization.
\newblock {\em arXiv preprint arXiv:1711.05101}, 2017.

\bibitem{lu2022dpm}
Cheng Lu, Yuhao Zhou, Fan Bao, Jianfei Chen, Chongxuan Li, and Jun Zhu.
\newblock Dpm-solver: A fast ode solver for diffusion probabilistic model
  sampling in around 10 steps.
\newblock {\em arXiv preprint arXiv:2206.00927}, 2022.

\bibitem{Midjourney}
\url{https://www.midjourney.com/home/?callbackUrl=%2Fapp%2F}, 2023.

\bibitem{muennighoff2022crosslingual}
Niklas Muennighoff, Thomas Wang, Lintang Sutawika, Adam Roberts, Stella
  Biderman, Teven~Le Scao, M~Saiful Bari, Sheng Shen, Zheng-Xin Yong, Hailey
  Schoelkopf, et~al.
\newblock Crosslingual generalization through multitask finetuning.
\newblock {\em arXiv preprint arXiv:2211.01786}, 2022.

\bibitem{nguyen2022empirical}
Nhan Nguyen and Sarah Nadi.
\newblock An empirical evaluation of github copilot's code suggestions.
\newblock In {\em Proceedings of the 19th International Conference on Mining
  Software Repositories}, pages 1--5, 2022.

\bibitem{openai2023gpt4}
OpenAI.
\newblock Gpt-4 technical report, 2023.

\bibitem{oppenlaender2022prompt}
Jonas Oppenlaender.
\newblock Prompt engineering for text-based generative art.
\newblock {\em arXiv preprint arXiv:2204.13988}, 2022.

\bibitem{ouyang2022training}
Long Ouyang, Jeffrey Wu, Xu~Jiang, Diogo Almeida, Carroll Wainwright, Pamela
  Mishkin, Chong Zhang, Sandhini Agarwal, Katarina Slama, Alex Ray, et~al.
\newblock Training language models to follow instructions with human feedback.
\newblock {\em Advances in Neural Information Processing Systems},
  35:27730--27744, 2022.

\bibitem{pennington2014glove}
Jeffrey Pennington, Richard Socher, and Christopher~D Manning.
\newblock Glove: Global vectors for word representation.
\newblock In {\em Proceedings of the 2014 conference on empirical methods in
  natural language processing (EMNLP)}, pages 1532--1543, 2014.

\bibitem{pressmancrowson2022}
John~David Pressman, Katherine Crowson, and Simulacra~Captions Contributors.
\newblock Simulacra aesthetic captions.
\newblock Technical Report Version 1.0, Stability AI, 2022.
\newblock \ url { https://github.com/JD-P/simulacra-aesthetic-captions }.

\bibitem{prolific}
\url{https://www.prolific.co}, 2023.

\bibitem{qadir2023engineering}
Junaid Qadir.
\newblock Engineering education in the era of chatgpt: Promise and pitfalls of
  generative ai for education.
\newblock In {\em 2023 IEEE Global Engineering Education Conference (EDUCON)},
  pages 1--9. IEEE, 2023.

\bibitem{radford2021learning}
Alec Radford, Jong~Wook Kim, Chris Hallacy, Aditya Ramesh, Gabriel Goh,
  Sandhini Agarwal, Girish Sastry, Amanda Askell, Pamela Mishkin, Jack Clark,
  et~al.
\newblock Learning transferable visual models from natural language
  supervision.
\newblock In {\em International conference on machine learning}, pages
  8748--8763. PMLR, 2021.

\bibitem{ramesh2022hierarchical}
Aditya Ramesh, Prafulla Dhariwal, Alex Nichol, Casey Chu, and Mark Chen.
\newblock Hierarchical text-conditional image generation with clip latents.
\newblock {\em arXiv preprint arXiv:2204.06125}, 1(2):3, 2022.

\bibitem{Rombach_2022_CVPR}
Robin Rombach, Andreas Blattmann, Dominik Lorenz, Patrick Esser, and Bj\"orn
  Ommer.
\newblock High-resolution image synthesis with latent diffusion models.
\newblock In {\em Proceedings of the IEEE/CVF Conference on Computer Vision and
  Pattern Recognition (CVPR)}, pages 10684--10695, June 2022.

\bibitem{rombach2022high}
Robin Rombach, Andreas Blattmann, Dominik Lorenz, Patrick Esser, and Bj{\"o}rn
  Ommer.
\newblock High-resolution image synthesis with latent diffusion models.
\newblock In {\em Proceedings of the IEEE/CVF Conference on Computer Vision and
  Pattern Recognition}, pages 10684--10695, 2022.

\bibitem{huggingface_dataset}
Gustavo Santana.
\newblock Stable-diffusion-prompts.
\newblock Huggingface Datasets, 2022.

\bibitem{schuhmann2022laionb}
Christoph Schuhmann, Romain Beaumont, Richard Vencu, Cade~W Gordon, Ross
  Wightman, Mehdi Cherti, Theo Coombes, Aarush Katta, Clayton Mullis, Mitchell
  Wortsman, Patrick Schramowski, Srivatsa~R Kundurthy, Katherine Crowson,
  Ludwig Schmidt, Robert Kaczmarczyk, and Jenia Jitsev.
\newblock {LAION}-5b: An open large-scale dataset for training next generation
  image-text models.
\newblock In {\em Thirty-sixth Conference on Neural Information Processing
  Systems Datasets and Benchmarks Track}, 2022.

\bibitem{simonyan2014very}
Karen Simonyan and Andrew Zisserman.
\newblock Very deep convolutional networks for large-scale image recognition.
\newblock {\em arXiv preprint arXiv:1409.1556}, 2014.

\bibitem{reuters_article_2023}
Karen Sloan.
\newblock A lawyer used chatgpt to cite bogus cases. what are the ethics?
\newblock {\em Reuters}, May 2023.

\bibitem{team2023gemini}
Gemini Team, Rohan Anil, Sebastian Borgeaud, Yonghui Wu, Jean-Baptiste Alayrac,
  Jiahui Yu, Radu Soricut, Johan Schalkwyk, Andrew~M Dai, Anja Hauth, et~al.
\newblock Gemini: a family of highly capable multimodal models.
\newblock {\em arXiv preprint arXiv:2312.11805}, 2023.

\bibitem{wang2022diffusiondb}
Zijie~J Wang, Evan Montoya, David Munechika, Haoyang Yang, Benjamin Hoover, and
  Duen~Horng Chau.
\newblock Diffusiondb: A large-scale prompt gallery dataset for text-to-image
  generative models.
\newblock {\em arXiv preprint arXiv:2210.14896}, 2022.

\bibitem{wei2022chain}
Jason Wei, Xuezhi Wang, Dale Schuurmans, Maarten Bosma, Ed~Chi, Quoc Le, and
  Denny Zhou.
\newblock Chain of thought prompting elicits reasoning in large language
  models.
\newblock {\em arXiv preprint arXiv:2201.11903}, 2022.

\bibitem{white2023prompt}
Jules White, Quchen Fu, Sam Hays, Michael Sandborn, Carlos Olea, Henry Gilbert,
  Ashraf Elnashar, Jesse Spencer-Smith, and Douglas~C Schmidt.
\newblock A prompt pattern catalog to enhance prompt engineering with chatgpt.
\newblock {\em arXiv preprint arXiv:2302.11382}, 2023.

\bibitem{wu2022promptchainer}
Tongshuang Wu, Ellen Jiang, Aaron Donsbach, Jeff Gray, Alejandra Molina,
  Michael Terry, and Carrie~J Cai.
\newblock Promptchainer: Chaining large language model prompts through visual
  programming.
\newblock In {\em CHI Conference on Human Factors in Computing Systems Extended
  Abstracts}, pages 1--10, 2022.

\bibitem{wu2022ai}
Tongshuang Wu, Michael Terry, and Carrie~Jun Cai.
\newblock Ai chains: Transparent and controllable human-ai interaction by
  chaining large language model prompts.
\newblock In {\em Proceedings of the 2022 CHI Conference on Human Factors in
  Computing Systems}, pages 1--22, 2022.

\bibitem{wu2023alsd}
Xiaoshi Wu, Keqiang Sun, Feng Zhu, Rui Zhao, and Hongsheng Li.
\newblock Better aligning text-to-image models with human preference.
\newblock {\em ArXiv}, abs/2303.14420, 2023.

\bibitem{xu2023imagereward}
Jiazheng Xu, Xiao Liu, Yuchen Wu, Yuxuan Tong, Qinkai Li, Ming Ding, Jie Tang,
  and Yuxiao Dong.
\newblock Imagereward: Learning and evaluating human preferences for
  text-to-image generation, 2023.

\bibitem{zhang2018unreasonable}
Richard Zhang, Phillip Isola, Alexei~A Efros, Eli Shechtman, and Oliver Wang.
\newblock The unreasonable effectiveness of deep features as a perceptual
  metric.
\newblock In {\em Proceedings of the IEEE conference on computer vision and
  pattern recognition}, pages 586--595, 2018.

\bibitem{zhou2022learning}
Kaiyang Zhou, Jingkang Yang, Chen~Change Loy, and Ziwei Liu.
\newblock Learning to prompt for vision-language models.
\newblock {\em International Journal of Computer Vision}, 130(9):2337--2348,
  2022.

\bibitem{zhou2022large}
Yongchao Zhou, Andrei~Ioan Muresanu, Ziwen Han, Keiran Paster, Silviu Pitis,
  Harris Chan, and Jimmy Ba.
\newblock Large language models are human-level prompt engineers.
\newblock {\em arXiv preprint arXiv:2211.01910}, 2022.

\bibitem{zhu2023collaborative}
Wanrong Zhu, Xinyi Wang, Yujie Lu, Tsu-Jui Fu, Xin~Eric Wang, Miguel Eckstein,
  and William~Yang Wang.
\newblock Collaborative generative ai: Integrating gpt-k for efficient editing
  in text-to-image generation.
\newblock {\em arXiv preprint arXiv:2305.11317}, 2023.

\end{thebibliography}

\appendix
\onecolumn

\section{Appendix}

\subsection{Dataset Limitations}
A potential limitation in our dataset is the diversity of unique images. However, as shown in our analysis (Sections~\ref{sec:prompt_diversity}, \ref{sec:analysis}), we have multiple illuminating findings despite having <200 unique images in our dataset. In collecting this dataset, there was a tradeoff – given a fixed budget to collect data, we could choose to collect more unique images or collect more user interactions per image. We opted to collect data on a fewer number of unique images with more users interacting with each image, as we believed this data would contain more insights into human interaction with the AI.

\subsection{Information on Wikipedia pages scraped}\label{app:wikipedia_pages}
Table~\ref{app:tbl:wikipedia_images} presents a list of the Wikipedia pages used to select real-world target images (see 
Section~\ref{sec:target_image_selection}). We extracted images from each listed Wikipedia page. We then uniformly sample a category and subsequently sample an image from a page in that category. This ensures a diverse set of images, which is important given that some of the Wikipedia pages contain many more images than others (e.g., \texttt{Paris} has 10 times more usable images than \texttt{Social\_documentary\_photography}).

\begin{table*}[ht]
\caption{Wikipedia images used.} 
\label{app:tbl:wikipedia_images}
\vskip 0.1in
\begin{center}
\begin{tabular}{|>{\centering\arraybackslash}m{1.2cm}|>{\centering\arraybackslash}m{8cm}|}
\toprule
\small Category & \small Wikipedia Pages \\ 
\hline\small Art & \small \texttt{Art}, \texttt{Fine\_art}, \texttt{Fine-art\_photography}, \texttt{History\_of\_art}, \texttt{Painting} \\
\hline
\small Astro & \small \texttt{Astrophotography} \\
\hline
\small Buildings & \small \texttt{Architectural\_photography}, \texttt{Architecture}, \texttt{Real\_estate} \\
\hline
\small City & \small \texttt{New\_York\_City}, \texttt{Paris}, \texttt{San\_Francisco}, \texttt{Seoul} \\
\hline
\small Fashion & \small \texttt{Fashion\_design}, \texttt{Fashion\_photography}, \texttt{Model\_(person)} \\
\hline
\small General & \small \texttt{Aerial\_photography}, \texttt{Culture}, \texttt{Documentary\_photography}, \texttt{Photography}, \texttt{Social\_documentary\_photography} \\
\hline
\small Landscape & \small \texttt{Landscape\_photography} \\
\hline
\small Nature & \small \texttt{Nature\_photography} \\
\hline
\small Plants & \small \texttt{Flower} \\
\hline
\small Portrait & \small \texttt{Mug\_shot}, \texttt{Portrait\_photography}, \texttt{Selfie} \\
\hline
\small US & \small \texttt{Americans}, \texttt{President\_of\_the\_United\_States}, \texttt{United\_States} \\
\hline
\small Wildlife & \small \texttt{Aquatic\_ecosystem}, \texttt{Macro\_photography}, \texttt{Marine\_habitats}, \texttt{Wildlife\_observation}, \texttt{Wildlife\_photography} \\

\bottomrule
\end{tabular} 
\end{center}
\vskip -0.1in
\end{table*}

\subsection{Examples of images generated for target parameter selection}\label{app:sec:target_image_examples}
\begin{figure*}
    \centering
    \includegraphics[width=5.5in]{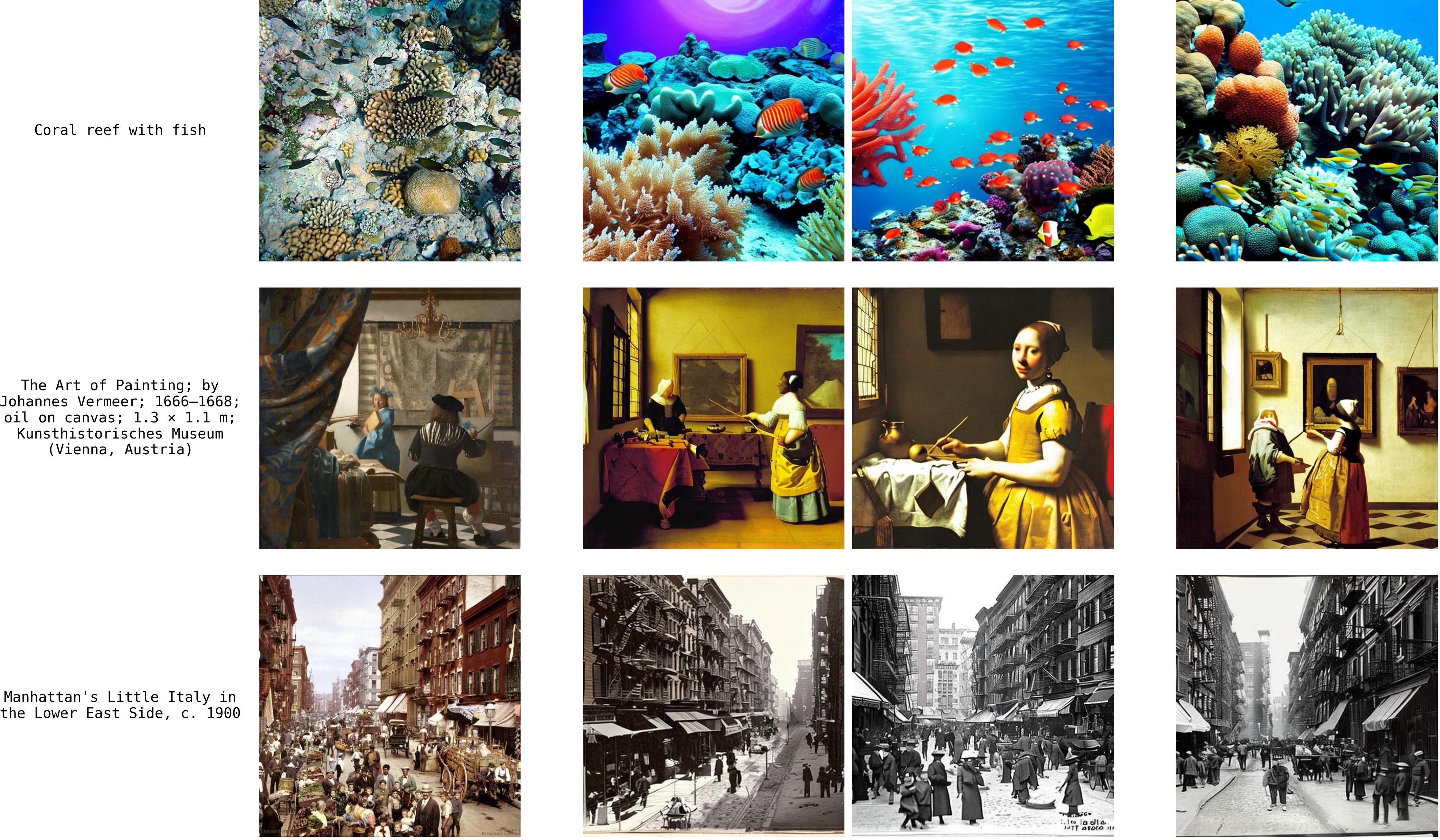}
    \caption{Wikipedia image examples generated using the target prompt (for target selection). From left to right: (1) original Wikipedia captions (the target prompt), (2) original photograph, (3-4) images generated with random seeds, (5) image generated using the fixed seed provided to users.}
    \label{app:fig:wikipedia_target_images}
\end{figure*}

\begin{figure*}
    \centering
    \includegraphics[width=5.5in]{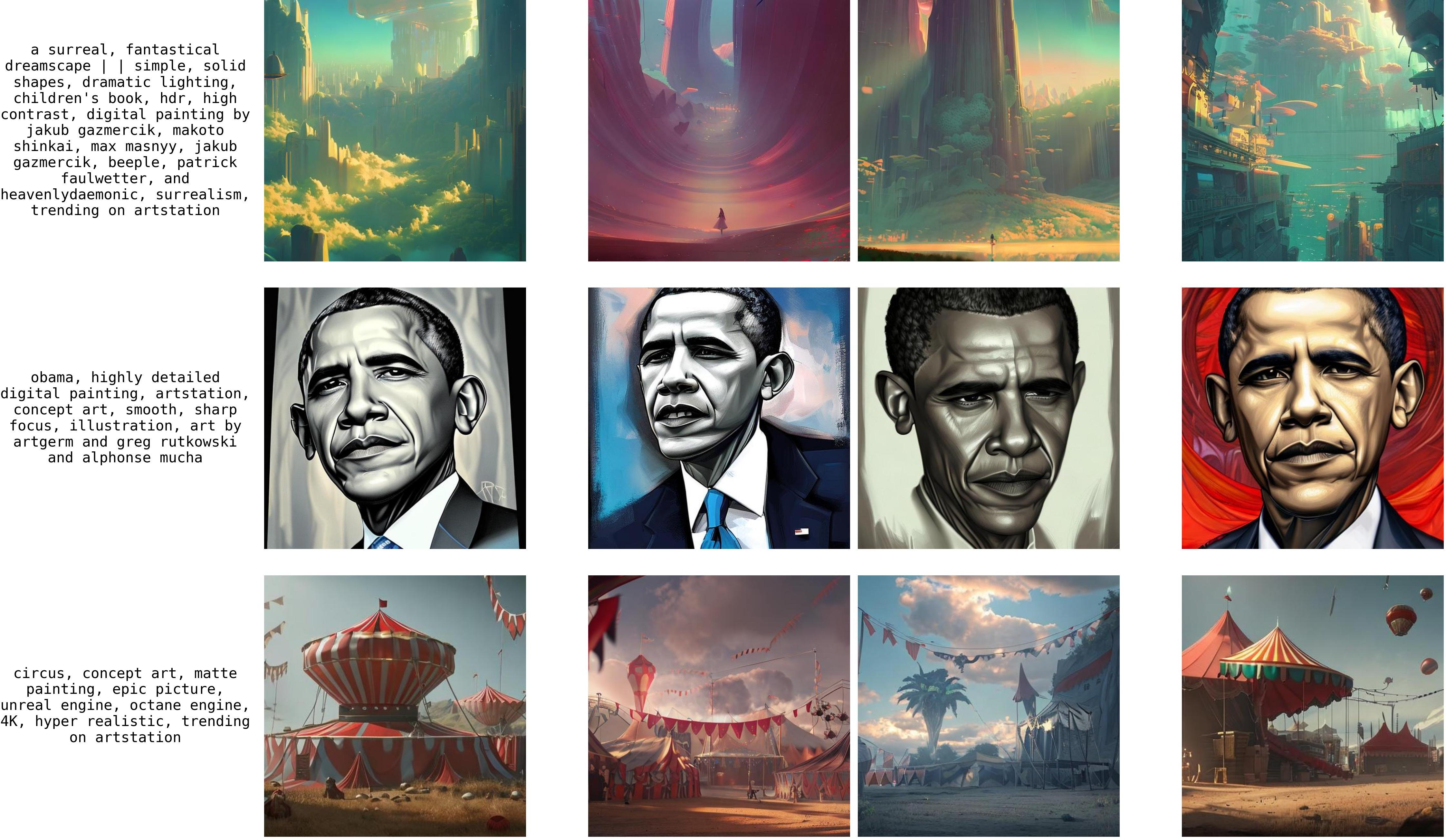}
    \caption{AI-Generated image examples on the target prompt (for target selection). From left to right: (1) target prompt, (2) target image (uses a different seed than the user's seed), (3-4) images generated with random seeds, (5) image generated using the fixed seed provided to users.}
    \label{app:fig:ai_target_images}
\end{figure*}

Here we provide a few examples of images generated during target parameter selection (see Section~\ref{sec:target_image_selection}). In Figure~\ref{app:fig:wikipedia_target_images}, we show the original (real) photograph (this is the target image shown to users), some examples of generated images using the caption as a prompt (i.e., the hidden goal for users), and the generated image using the selected parameters with the caption as a prompt (i.e., the generated image when a user finds the ``best'' prompt using the selected seed, $s_{i^*}$, from Equation~\ref{eqn:wiki_seed}). In Figure~\ref{app:fig:ai_target_images}, we show the target image shown to users ($t_{i^*}$ from Equation~\ref{eqn:ai_target_image}), examples of generated images using the caption on various random seeds, as well as the image generated when using the seed provided to users (i.e., using the seed $s_{i_2^*}$). For both Figures, scores are normalized with respect to the rightmost image in each row -- this image is guaranteed to score $100$ / $100$. Notice that the images generated from other seeds (the center two images) are also similar in quality to the image generated from the selected seed (the rightmost image).

\subsection{Scoring function details}\label{app:scoring_function}
In Section~\ref{sec:scoring_function}, we defined the scoring function, $score(x,t)$ as 
        $$ score(x, t) = \max(0, \min(100, \alpha \cdot \left|\left| \frac{C(x)}{||C(x)||_2} - \frac{C(t)}{||C(t)||_2} \right|\right|_2 + \beta)), $$
$\alpha$ and $\beta$ are constants used to scale the embedding distance prior to clipping the score. To select $\alpha$ and $\beta$, we used a small dataset of interactions collected by the authors prior to the main dataset collection (this data is not included in the released dataset). This dataset contains groups of images paired with scores in the range $[0,1]$. For each target image in this small dataset (5 in total), we add the following images to the dataset:
\begin{enumerate}
    \item AI-generated images that use the target prompt but with a different seed. These images are assigned a score of $1$.
    \item The images corresponding to the human-generated prompts. These images are assigned a score of $0.5$.
    \item AI-generated images that use a different prompt than the target prompt. These images are assigned a score of $0$.
\end{enumerate}
The intuition here is that with the AI-generated images, we can assume using the same target prompt with a different seed should generate a similar image hence the highest score possible ($1$). Using a different prompt (from our prompt dataset \cite{huggingface_dataset}), however, should result in an entirely different image hence the lowest score possible ($0$). Images generated by people are assumed to be somewhere in between, hence the score of $0.5$.

We then fit a linear regression model to this dataset (to predict score given the CLIP image embedding), using balanced sampling across the image groups. This linear model has parameters
\begin{align*}
    \alpha' &= -1.503 \\
    \beta' &= 1.791
\end{align*}
Since our score range is $[0,100]$, we scale the model parameters by $100$, resulting in
\begin{align*}
    \alpha'' &= -150.3 \\
    \beta'' &= 179.1
\end{align*}

We also add a ``score adjustment'' term that attempts to normalize image difficulty. In particular, for each target image, we compute the un-clipped score for the target image $t_k$ and the image generated using the target prompt with, $x_k$:
    $$ unclipped\_score(x_k, t_k) = \alpha'' \cdot \left|\left| \frac{C(x_k)}{||C(x_k)||_2} - \frac{C(t_k)}{||C(t_k)||_2} \right|\right|_2 + \beta'' $$
This score assess the score a user would receive if they exactly entered the target prompt. We fix this value as $100$ (i.e., a perfect score prior to clipping), and set the score adjustment parameter, $c_k$, to appropriately normalize this score. In particular,
    $$ c_k = \frac{100}{unclipped\_score(x_k,t_k)}, $$
and then we obtain target specific parameters,
\begin{align*}
    \alpha_k &= -150.3 \cdot c_k \\
    \beta_k &= 179.1 \cdot c_k
\end{align*}

Note that this means each target image may have slightly different parameters as the $c_k$ values vary across images.

\subsection{Additional information on running the game}

\paragraph{Why we limit user input?}
We deliberately limited user input to only a prompt, as opposed to giving users access to the random seed or other model hyperparameters. This was done for a few reasons:
\begin{enumerate}
    \item[1.] We wanted all users who enter the same prompt for a given image to see the same output.
    \item[2.] We wanted to limit the complexity of the task for users less/unfamiliar with text-to-image models.
    \item[3.] We wanted users to generate new prompts and not just resample new seeds until getting lucky. While users could still employ a version of this “random resampling” strategy by making small changes to their prompts, we did not want to encourage this practice through a seed parameter. This random resampling strategy, while useful in practice, is not such an interesting result for research purposes as it is easy to simulate random resampling strategies without any user input.
\end{enumerate}

\paragraph{Additional Technical Details}\label{sec:game_technical_details}
For the generative model, we use SD v2.1 \cite{rombach2022high} with the DPM Multi-step Scheduler \cite{lu2022dpm} and run the model for 20 iterations. AI-generated target images use the same parameters but run for 50 iterations. 20 iterations was selected to limit latency for players to 1-3 seconds depending on the player's internet connection. All images are generated at size 512 $\times$ 512.

\begin{figure*}
    \centering
    \includegraphics[width=5.5in]{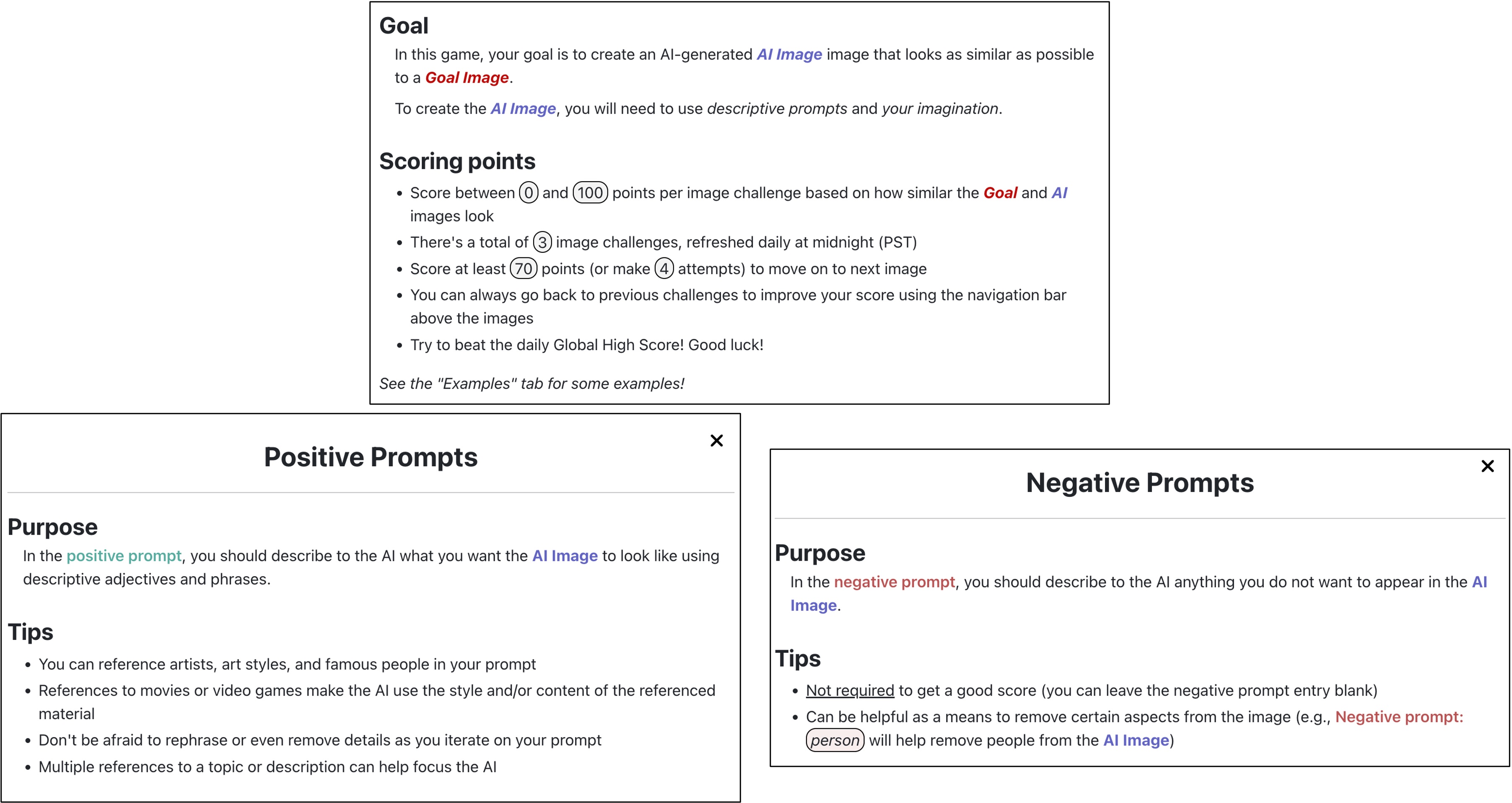}
    \caption{Game instructions}
    \label{app:fig:game_instructions}
\end{figure*}

\paragraph{Game instructions}
Game instructions are provided in Figure~\ref{app:fig:game_instructions}. Here we show the main instructions provided on how to play (top), as well as the tool-tips given for positive prompts (lower left) and negative prompts (lower right).

\paragraph {Crowd workers}
Crowd workers are adults from the US. They were paid at a rate of $\$12.00$ per hour for roughly 20 minutes of time. Additionally, they were provided bonus payment of between $\$0.10-\$0.50$ per image they received a perfect score on (depending on the image difficulty). In total, we paid about $600$ for recruiting the crowd workers.

\subsection{Additional Stats}

\begin{figure*}
    \centering
    \includegraphics[width=4in]{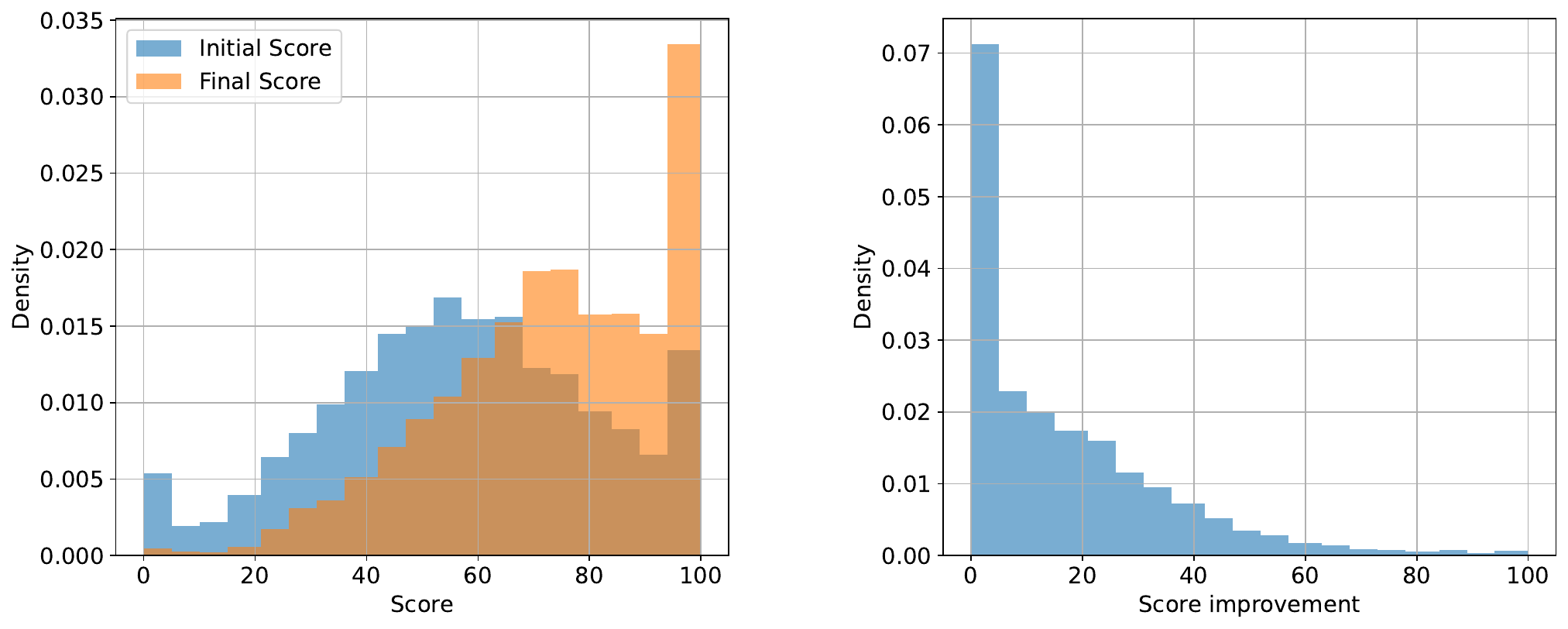}
    \caption{On the left: distribution of initial and final scores (across users and targets). Average initial score is 56.7; average final score is 73.0. On the right: distribution of image improvement (difference between final and initial score). Average score improvement is 16.3.}
    \label{app:fig:score_distribution}
\end{figure*}

In Figure~\ref{app:fig:score_distribution}, we plot the distribution of user scores across target images. The left plot shows the initial and final scores. The shift in score to the right indicates the improvement in generated image similarity to target image. The average initial score is 56.7 (median is 57.0), and the average final score is 73.0 (median is 75.0). On the left, we plot the distribution of score improvement (the difference between the final and initial scores). The  large density of 0 or close to 0 improvement is due to the ``easiest'' target images that users were able to generated similar examples of on their initial attempt (for example, see the first two rows of Figure~\ref{app:fig:steerability_for_individual_images}). Most interactions result in a score difference of at most 50 points, as most users score above 50 points on their initial attempt, limiting the amount of improvement possible (the median initial score is 57.0). 

\begin{figure*}
    \centering
    \includegraphics[width=4in]{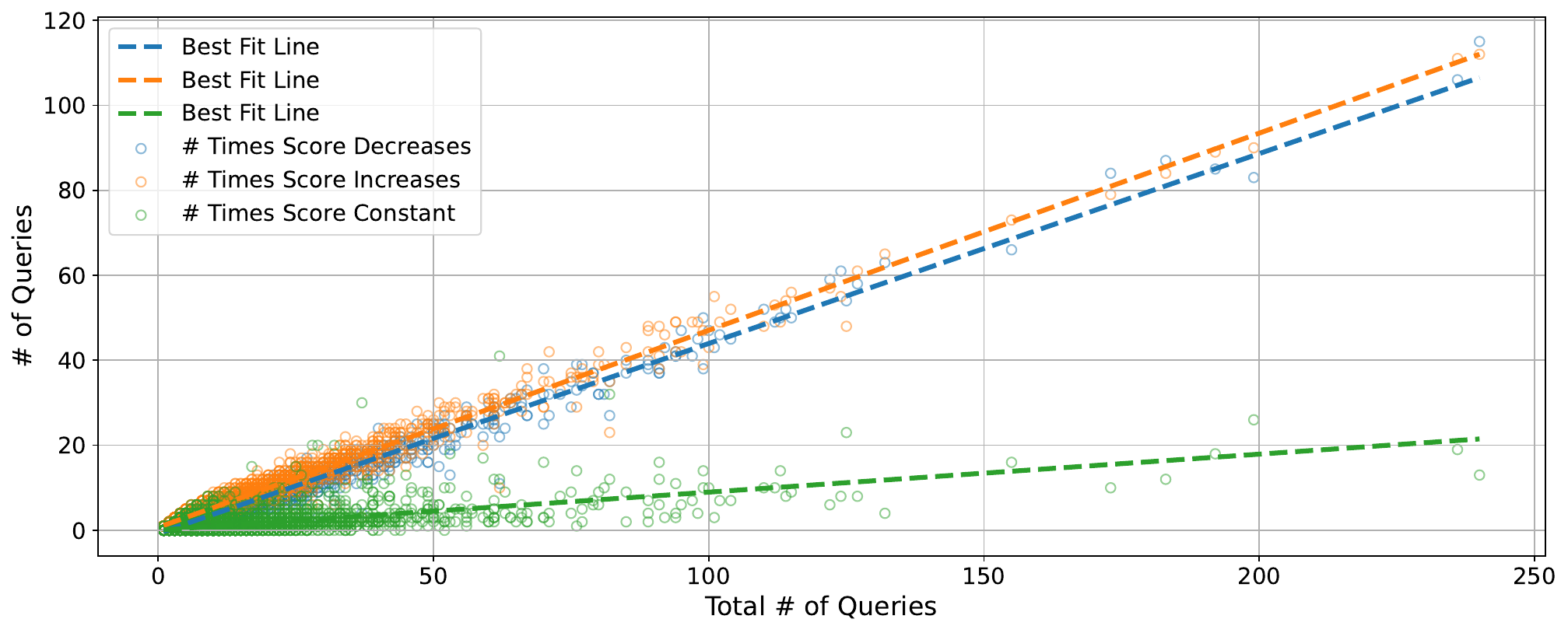}
    \caption{How score changes between queries. Each circle represents a series of user interactions to a generate a single target image. In blue, we plot the number of times a user's score decreases over that interaction (between two queries); similarly, in orange and green we plot the number of times the user's score increases or stays constant between prompts. We also plot best fit lines (fit using linear least squares).}
    \label{app:fig:score_changes}
\end{figure*}

In Figure~\ref{app:fig:score_changes}, we plot how often a user's score decreases, increases, or remains constant between prompts. Each point represents a series of user interactions to a generate a single target image. We note that the ratio of score increases to score decreases is similar, with the number of score increases being slightly higher. This indicates that score improvement (and so, an increase in the similarity of the generated to target image) may be effectively represented by a stochastic process with a slight bias towards score increases. This observation is the basis on which we construct our proposed steerability metric (i.e., representing score change as a stochastic process).

\subsection{ArtWhisperer-Validation Statistics}\label{app:validation_data_overview}
\begin{figure*}
    \centering
    \includegraphics[width=5.5in]{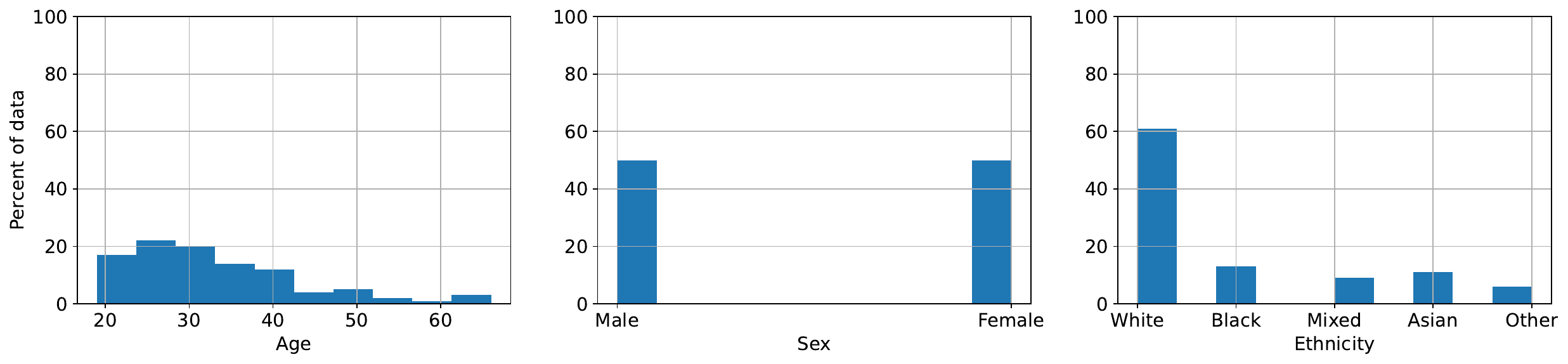}
    \caption{Demographic information for the paid crowd workers in the ArtWhisperer-Validation dataset.}
    \label{app:fig:validation_demographics}
\end{figure*}

In Table~\ref{app:tbl:overview_of_validation_data}, we provide general statistics about the \textit{ArtWhisperer-Validation} Dataset. This information mirrors that statistics provided in Table~\ref{tbl:overview_of_data} for the ArtWhisperer-Validation dataset. We also provide demographic information collected through Prolific in Figure~\ref{app:fig:validation_demographics}. Users were sampled through Prolific to guarantee an even split of male and female users.

\begin{table}[ht]
\caption{\textit{ArtWhisperer-Validation} Dataset Overview. Each row contains summary data for a different subset of the dataset. Subsets may overlap.
} 
\label{app:tbl:overview_of_validation_data}
\vskip -0.3in
\begin{center}
\begin{tabular}{>{\centering\arraybackslash}m{1.2cm}|>{\centering\arraybackslash}m{1.1cm}|>{\centering\arraybackslash}m{1.1cm}|>{\centering\arraybackslash}m{1.3cm}|>{\centering\arraybackslash}m{1.1cm}|>{\centering\arraybackslash}m{1.2cm}|>{\centering\arraybackslash}m{4cm}}
\toprule
\small \# Players & \small \# Target Images & \small \# Interactions & \small Average \# Prompts & \small Average Score & \small Median Duration & \small Category \\ 
\hline\hline
\small \textbf{140} & \small \textbf{51} & \small \textbf{4572} & \small \textbf{8.14} & \small \textbf{54.77} & \small \textbf{24 s} & \small \textbf{Total} \\
\hline\hline
\small 73 & \small 9 & \small 762 & \small 7.47 & \small 47.55 & \small 20 s & \small Famous person? \\
\hline
\small 134 & \small 42 & \small 3937 & \small 8.19 & \small 54.58 & \small 23 s & \small Manmade? \\
\hline
\small 124 & \small 26 & \small 2154 & \small 7.69 & \small 54.61 & \small 24 s & \small Real image? \\
\hline
\small 123 & \small 24 & \small 2424 & \small 9.32 & \small 50.96 & \small 24 s & \small Art? \\
\hline
\small 94 & \small 13 & \small 1044 & \small 7.05 & \small 57.93 & \small 25 s & \small Famous landmark? \\
\hline
\small 72 & \small 9 & \small 858 & \small 8.94 & \small 50.33 & \small 26 s & \small Nature? \\
\hline
\small 70 & \small 8 & \small 730 & \small 8.11 & \small 60.25 & \small 25 s & \small Sci-fi or space? \\
\hline
\small 94 & \small 16 & \small 1501 & \small 8.25 & \small 54.21 & \small 24 s & \small People? \\
\hline
\small 123 & \small 25 & \small 2418 & \small 8.57 & \small 54.90 & \small 24 s & \small AI image? \\
\hline
\small 75 & \small 9 & \small 786 & \small 7.42 & \small 59.11 & \small 23 s & \small City? \\
\hline
\small 75 & \small 9 & \small 808 & \small 8.16 & \small 57.56 & \small 26 s & \small Fantasy? \\

\bottomrule
\end{tabular} 
\end{center}
\vskip -0.1in
\end{table}

\subsection{Results for Alternative Text Embeddings}\label{app:alternative_text_embeddings}
\begin{figure*}
    \centering
    \includegraphics[width=5.5in]{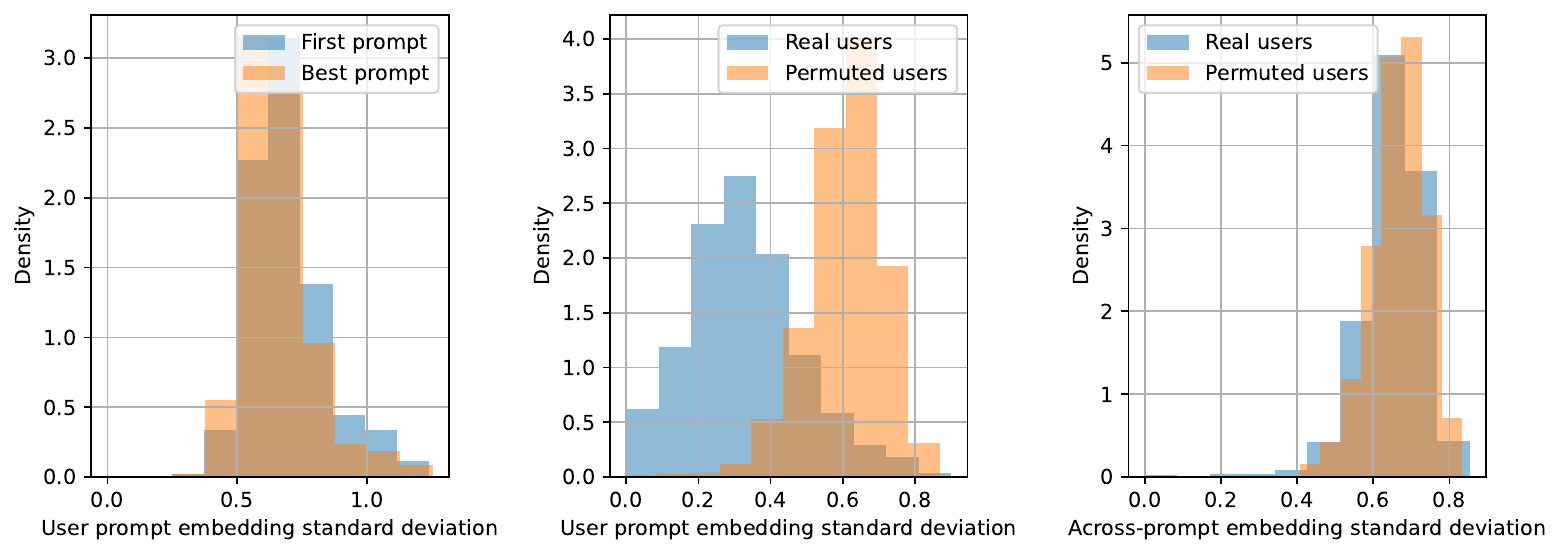}
    \caption{Same plot as Figure~\ref{fig:prompt_diversity} but using the BERT \cite{devlin2018bert} text embedding instead of CLIP.}
    \label{app:fig:prompt_diversity_bert}
\end{figure*}

In Figure~\ref{app:fig:prompt_diversity_bert}, we replicate the analysis from Figure~\ref{fig:prompt_diversity} but using a BERT \cite{devlin2018bert} embedding rather than a CLIP embedding. We note similar findings and significance to the CLIP embedding analysis. We also repeated this analysis using the GLOVE \cite{pennington2014glove} text embedding and again see similar results. This indicates that our analysis is robust across text representations.

\subsection{Additional example images}
\begin{figure*}
    \centering
    \includegraphics[width=5in]{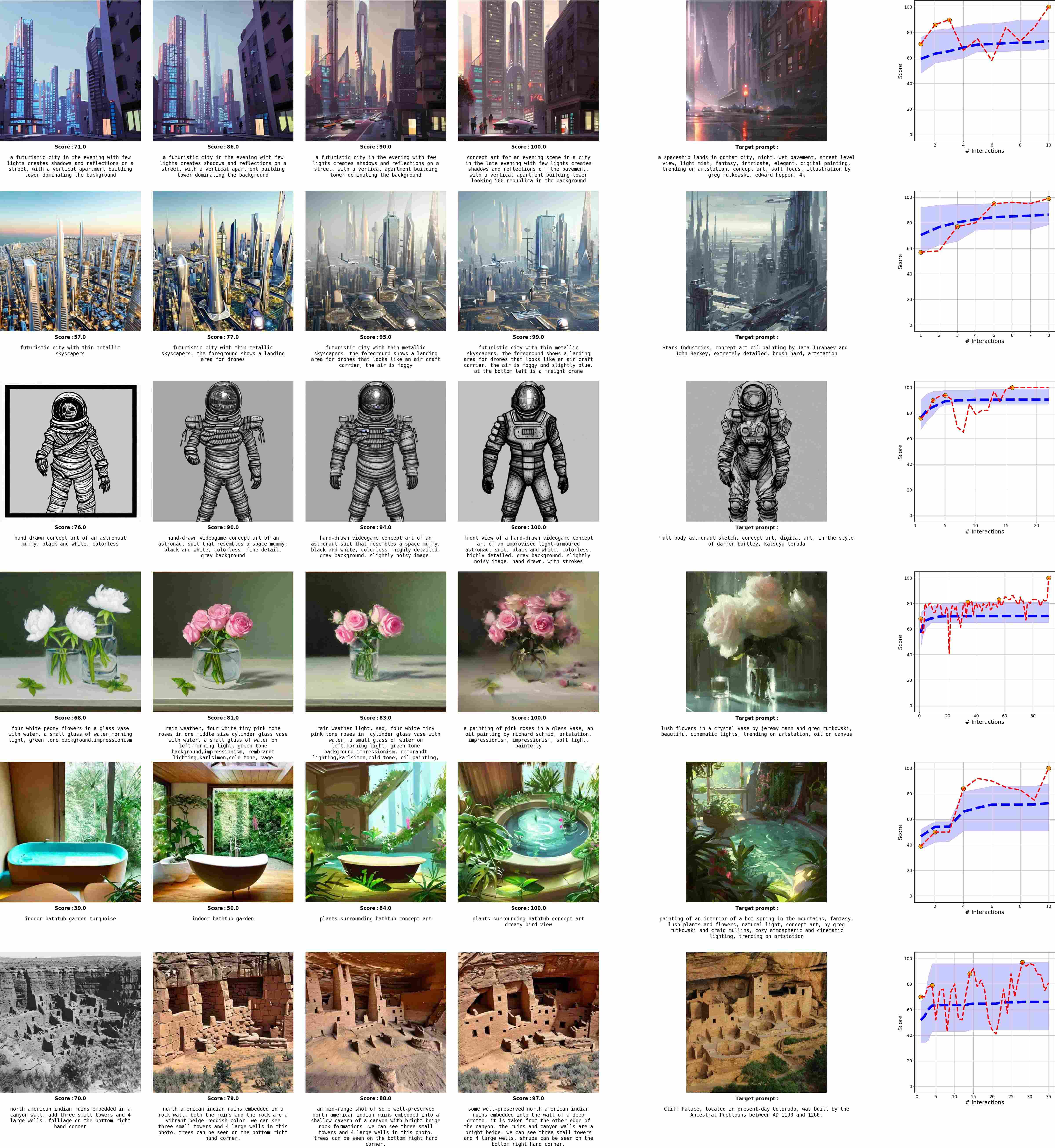}
    \caption{More examples of user trajectories, as in Figure~\ref{fig:ex_target_images}.}
    \label{app:fig:ex_target_images}
\end{figure*}

\begin{figure*}
    \centering
    \includegraphics[width=4.5in]{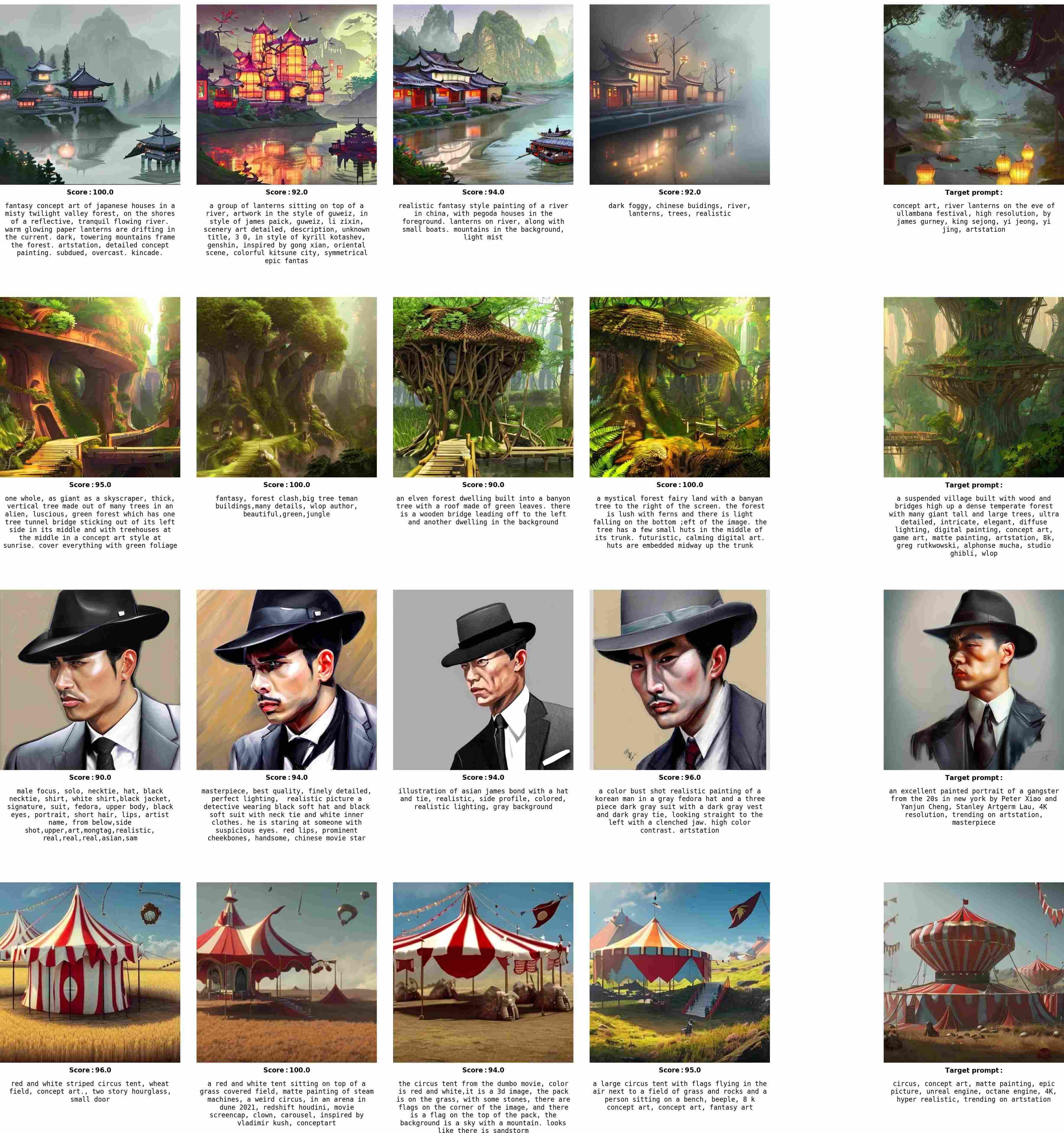}
    \caption{More examples of diverse, high-scoring prompts, as in Figure~\ref{fig:diverse_prompt_examples}.}
    \label{app:fig:diverse_prompt_examples}
\end{figure*}

We provide additional examples of image trajectories and diverse images in Figures~\ref{app:fig:ex_target_images} and \ref{app:fig:diverse_prompt_examples}.

\subsection{Algorithm for steerability}
We describe the algorithm for assessing steerability in more detail in Algorithm~\ref{app:alg:estimate_steerability}. Here, we define three procedures. \texttt{EstimateStoppingTime} estimates the steerability of a target image. We define a set of score bins; we chose 5 equally sized bins so that there was sufficient data to cover each bin. We also use a regularizer, $\epsilon$. What $\epsilon$ essentially does is encode a prior that from any given score, the transition to a new score is uniformly random. Then for each target image, we find the empirical score transition probabilities in \texttt{EstimateMarkovChain} and use Monte Carlo simulation to estimate the stopping time in \texttt{RunMonteCarloEstimation}.

\begin{algorithm}[tb]
    \caption{Image Steerability Estimation}
    \label{app:alg:estimate_steerability}
    \begin{algorithmic}[1]
        \STATE {\bfseries define function} \texttt{EstimateSteerability}
        \STATE \hspace{1em} {\bfseries Input:} set of images, $targetImages$
        \STATE \hspace{1em} Initialize array $steerability$ to track stopping times
        \STATE \hspace{1em} Initialize $scoreBins \leftarrow [[0,20],[21,40],[41,60],[61,80],[81,100]]$
        \STATE \hspace{1em} Initialize $\epsilon \leftarrow 1$
        \STATE \hspace{1em} {\bfseries for} ${image}_i \in \text{targetImages}$ {\bfseries do}
            \STATE \hspace{2em} ${markov}_i$ $\leftarrow$ \texttt{EstimateMarkovChain}(${image}_i$, $scoreBins$, $\epsilon$)
            \STATE \hspace{2em} $steerability[i]$ $\leftarrow$ \texttt{RunMonteCarloEstimation}(${markov}_i$)
        \STATE \hspace{1em} {\bfseries endfor}
        \STATE \hspace{1em} {\bfseries Return:} $\mathbf{E}[steerability]$
    \end{algorithmic}
    
    \hspace{1em}
    
    \begin{algorithmic}[1]
        \STATE {\bfseries define function} \texttt{EstimateMarkovChain}
        \STATE \hspace{1em} {\bfseries Input:} targetImage, bins, $\epsilon$
        \STATE \hspace{1em} Initialize bin pair count as $counts[(\text{bin}_i, \text{bin}_j)]\leftarrow \epsilon$, using transitions from a dummy node to model the first prompt submitted
        \STATE \hspace{1em} {\bfseries for} $\text{user}_i \leftarrow \text{user}_1$ {\bfseries to} $\text{user}_n$ {\bfseries do}
            \STATE \hspace{2em} {\bfseries for} $\text{score}_{i,j} \leftarrow \text{score}_{i,1}$ {\bfseries to} $\text{score}_{i,d}$ {\bfseries do}
                \STATE \hspace{3em} Convert $\text{score}_{i,j}$ to bin number, $\text{bin}_{i,j}$
                \STATE \hspace{3em} Increment $counts[(\text{bin}_{i-1,j}, \text{bin}_{i,j})]$ by $1$
            \STATE \hspace{2em} {\bfseries endfor}
        \STATE \hspace{1em} {\bfseries endfor}
        \STATE \hspace{1em} Normalize empirical transition counts to find empirical node transition probabilities
        \STATE \hspace{1em} Define $markov_{\text{targetImage}}$ using the node transition probabilities
        \STATE \hspace{1em} {\bfseries Return:} $markov_{\text{targetImage}}$
    \end{algorithmic}

    \hspace{1em}
    
    \begin{algorithmic}[1]
        \STATE {\bfseries define function} \texttt{RunMonteCarloEstimation}
        \STATE \hspace{1em} {\bfseries Input:} $markovChain$
        \STATE \hspace{1em} Run Monte Carlo simulation to estimate time to reach the last bin for $\text{markovChain}$, starting from the dummy (initial) node
        \STATE \hspace{1em} {\bfseries Return:} \textit{Estimated stopping time}
    \end{algorithmic}

\end{algorithm}

\subsection{Steerability across models}\label{app:steerability_across_models}

\begin{figure}
    \centering
    \includegraphics[width=5in]{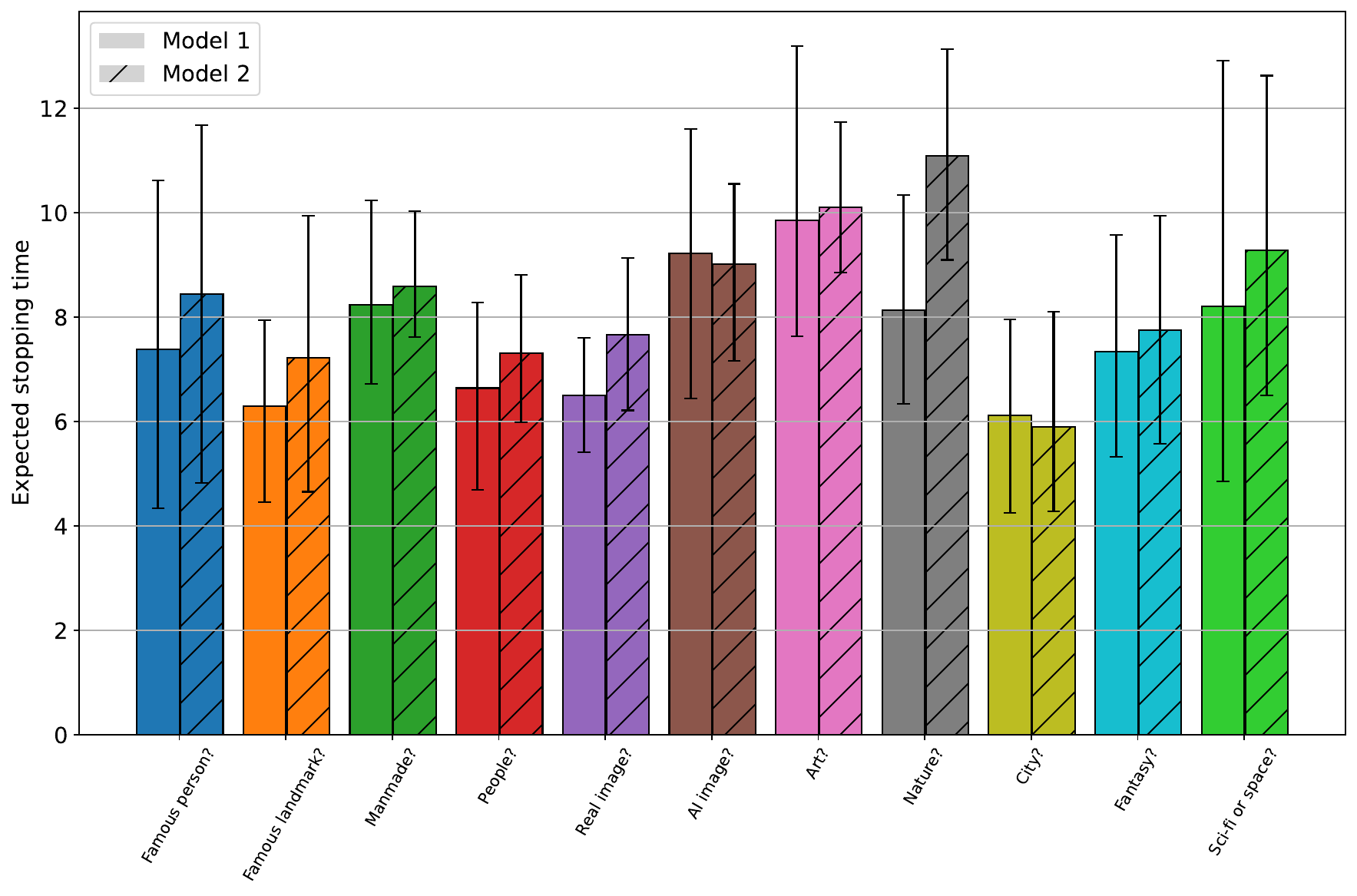}
    \caption{Steerability across image groups, comparing between models using a subset of the images. ''Model 1`` is Stable Diffusion v2.1 and ''Model 2`` is Stable Diffusion v1.5. Note only a subset of the images are used to compare the two models, hence the differences between ''Model 1`` and the results in Figure~\ref{fig:steerability_across_images}.}
    \label{app:fig:steerability_across_models}
\end{figure}

In Figure~\ref{app:fig:steerability_across_models}, we plot the steerability across SDv2.1 and SDv1.5. As described in Section~\ref{sec:analysis}, images of nature, sci-fi or space, and real images have the largest differences in steerability between the two models. ``Nature'' is the only image group with a steerability difference greater than the standard deviation of the mean. Other image groups seem to have similar performance across both SDv2.1 and SDv1.5. This suggests that SDv2.1 only makes minor improvement over SDv1.5 across most image categories.

\subsection{Steerability scores for individual images}\label{app:individual_image_steerability}
\begin{figure*}
    \centering
    \includegraphics[width=4in]{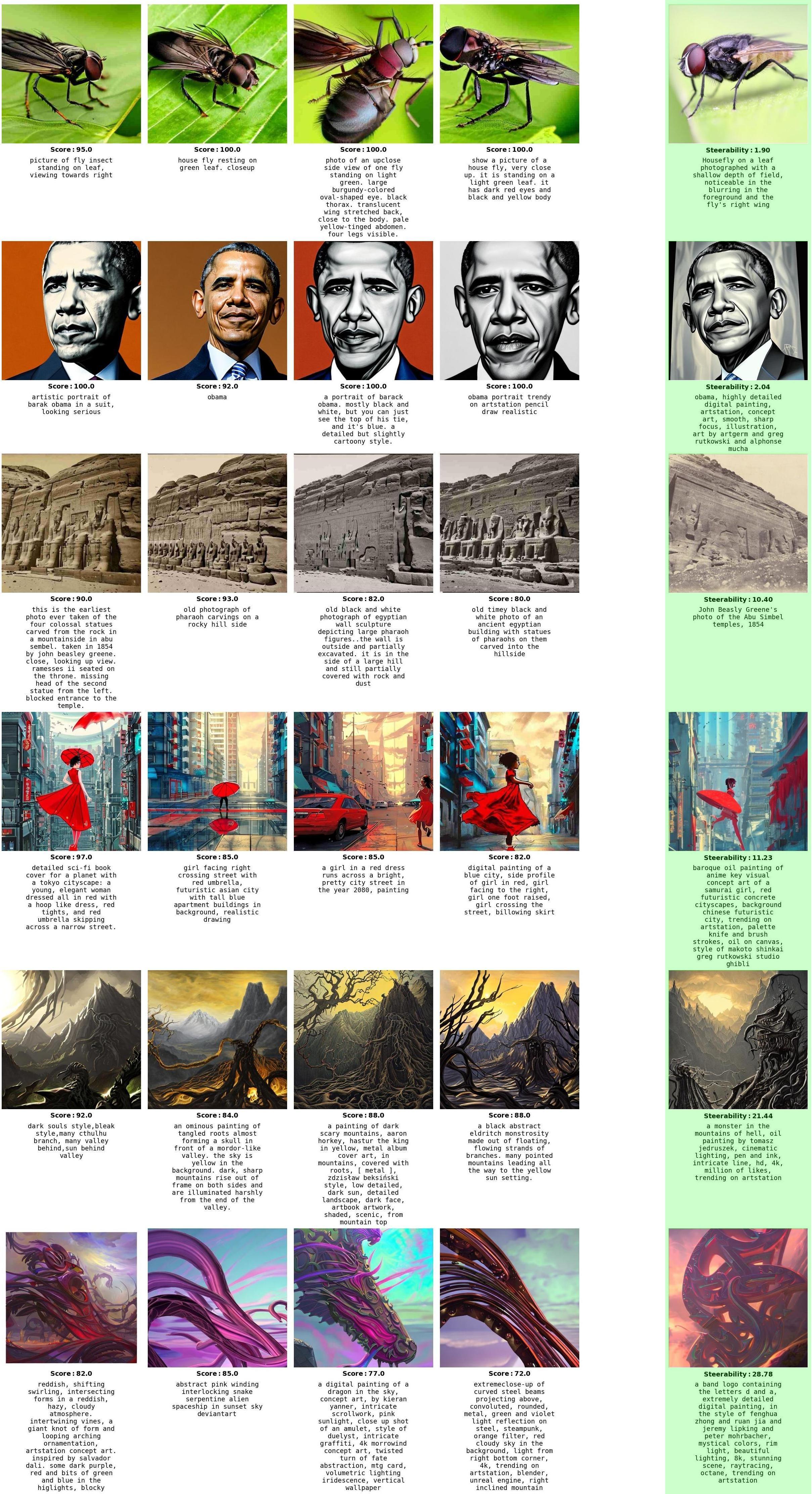}
    \caption{Steerability of individual target images. Each row is a different target image. The first four columns are examples of user submitted prompts with the corresponding AI-generated images. Rightmost column shows the target image along with the \emph{steerability} of that image. A smaller steerability value corresponds to the model being easier to steer for the given target image.}
    \label{app:fig:steerability_for_individual_images}
\end{figure*}

In Figure~\ref{app:fig:steerability_for_individual_images}, we provide some example images along with their steerability scores. Note that more simple images with well-defined content that likely has high presence in the model's training data (e.g., the first two rows--a fly on a leaf; a drawing of Barack Obama, well-known public figure) have smaller steerability values, indicating they are easier to steer. However, more complex content that is also more ambiguous for users (e.g., the last three rows), have larger steerability indicating greater difficulty in steering. 

\begin{figure*}
    \centering
    \includegraphics[width=4in]{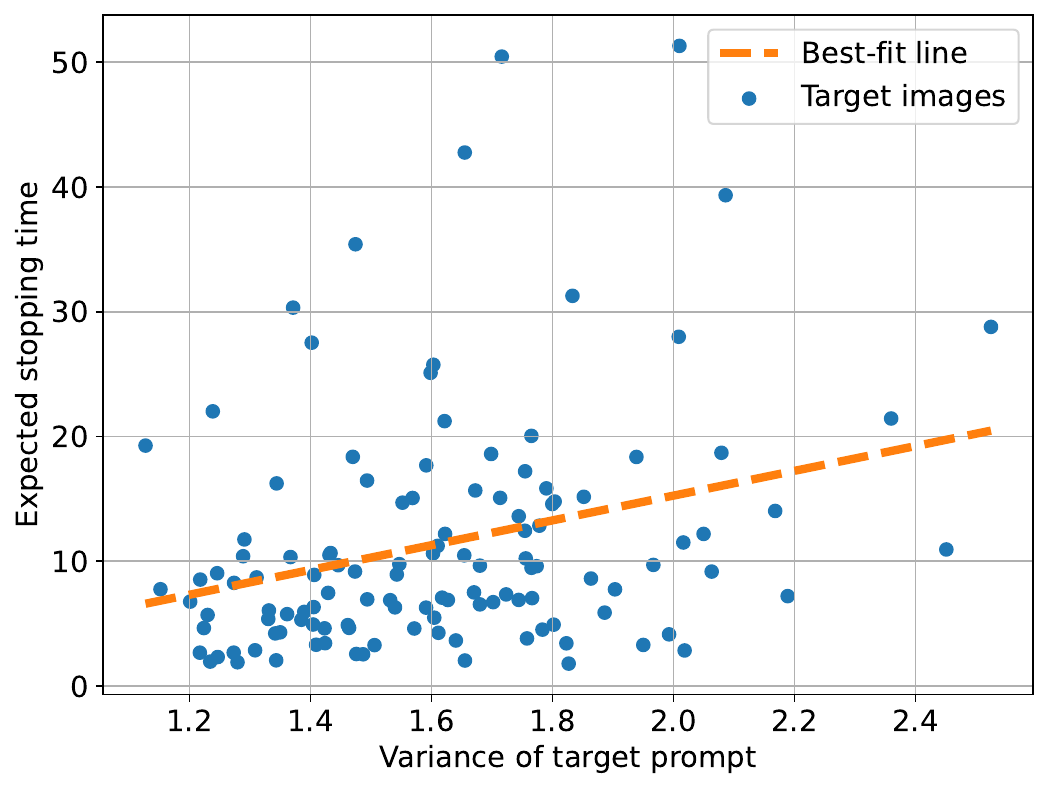}
    \caption{Steerability decreases (expected stopping time increases) when a model has higher variance given a target prompt.}
    \label{app:fig:steerability_vs_variance}
\end{figure*}

We also find that image steerability is negatively correlated with image variance (across seeds for a fixed prompt). Consider fixing a target prompt and sampling the model across many seeds. When the model outputs a wide variety of images, the variance of the output increases (i.e., computed in the image embedding space). We find that target images that have higher variance are also less steerable (indicated by a higher expected stopping time). In other words, if repeatedly sampling a model with the same prompt can produce a wide range of outputs (high variance), then the model is likely less steerable for the content in those generated images. We plot this result in Figure~\ref{app:fig:steerability_vs_variance}.

\subsection{Content knowledge increases steerability}\label{app:user_familiarity}
\begin{figure*}
    \centering
    \includegraphics[width=4in]{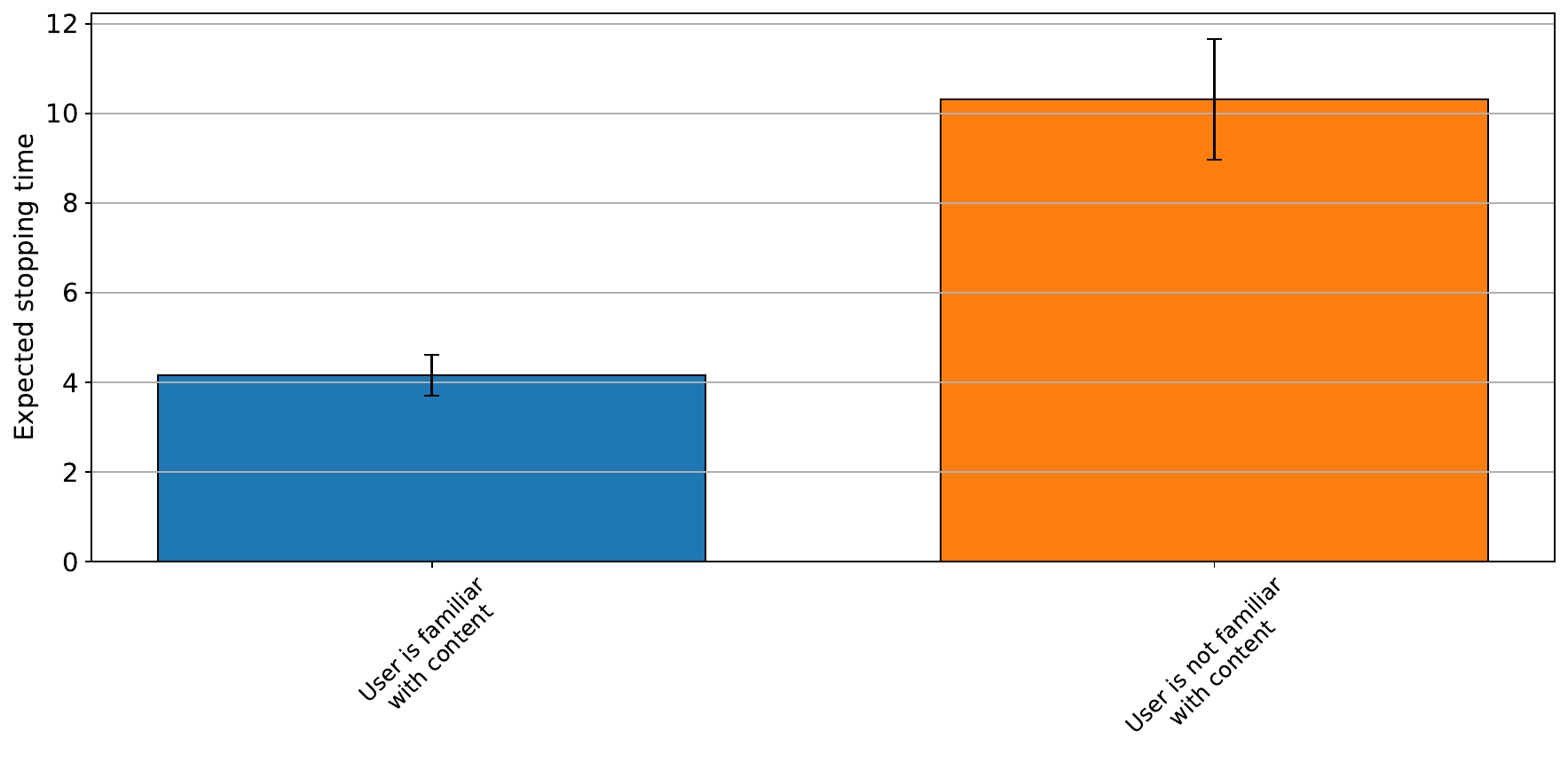}
    \caption{Steerability across familiarity with target image content. When a user is more familiar with the subject matter of the target image, it is significantly easier to steer the model.}
    \label{app:fig:user_familiarity}
\end{figure*}

Knowledge of an image's subject matter increases steerability.
In Figure~\ref{app:fig:user_familiarity}, we plot the steerability across two groups of individuals -- those who had subject matter knowledge and those would did not. To define this split, we only consider the subset of images that contain either a famous person or landmark. Users were defined to have knowledge of the subject matter when they referenced a key word (e.g., the famous person's name) in at least one of their submitted prompts. We then computed steerability across these two groups of users. As is indicated in the plot, users with subject matter knowledge found the model to be significantly more steerable.

\subsection{Additional discussion around human ratings}\label{app:human_rating_discussion}
Here we provide plots for the analysis using human ratings. In Figure~\ref{app:fig:human_rating_correlation}, we show a scatter plot of scores and human ratings. We also plot a best fit line which has a correlation of 0.597, indicating that our score function produces values that are indeed similar to the human ratings.

We also provide a plot comparing the steerability value calculated using our score function and calculated using the human ratings in Figure~\ref{app:fig:human_rating_vs_score}. Error bars indicate the standard error. Images depicting sci-fi or space have the greatest difference (humans seem to be harsher judges of their generated images' similarity to the target. However, for most image groups, the two steerability scores are quite close and generally exceed a 95\% confidence interval.

\subsubsection{On our choice of scoring metric and its relation to human ratings}
These results validate our scoring metric based on CLIP. As is well known, embeddings extracted from deep models are reasonable at assessing image similarity \cite{zhang2018unreasonable}. We chose to base our scoring metric using the CLIP embedding in particular, however, for two reasons: (1) its recent usage in the literature for effectively representing semantic meaning in images \cite{radford2021learning, khandelwal2022simple} and (2) after testing a number of methods on a small subset of trial data before launching our data collection, we found the CLIP embedding was comparable or better than any of the other methods including ensemble-based approaches. (The methods we tested include embeddings extracted from other deep learning architectures (including ResNet \cite{he2016deep} and VGG \cite{simonyan2014very}) as well as classical image embeddings (like color histograms).) 

\begin{figure*}
    \centering
    \includegraphics[width=3in]{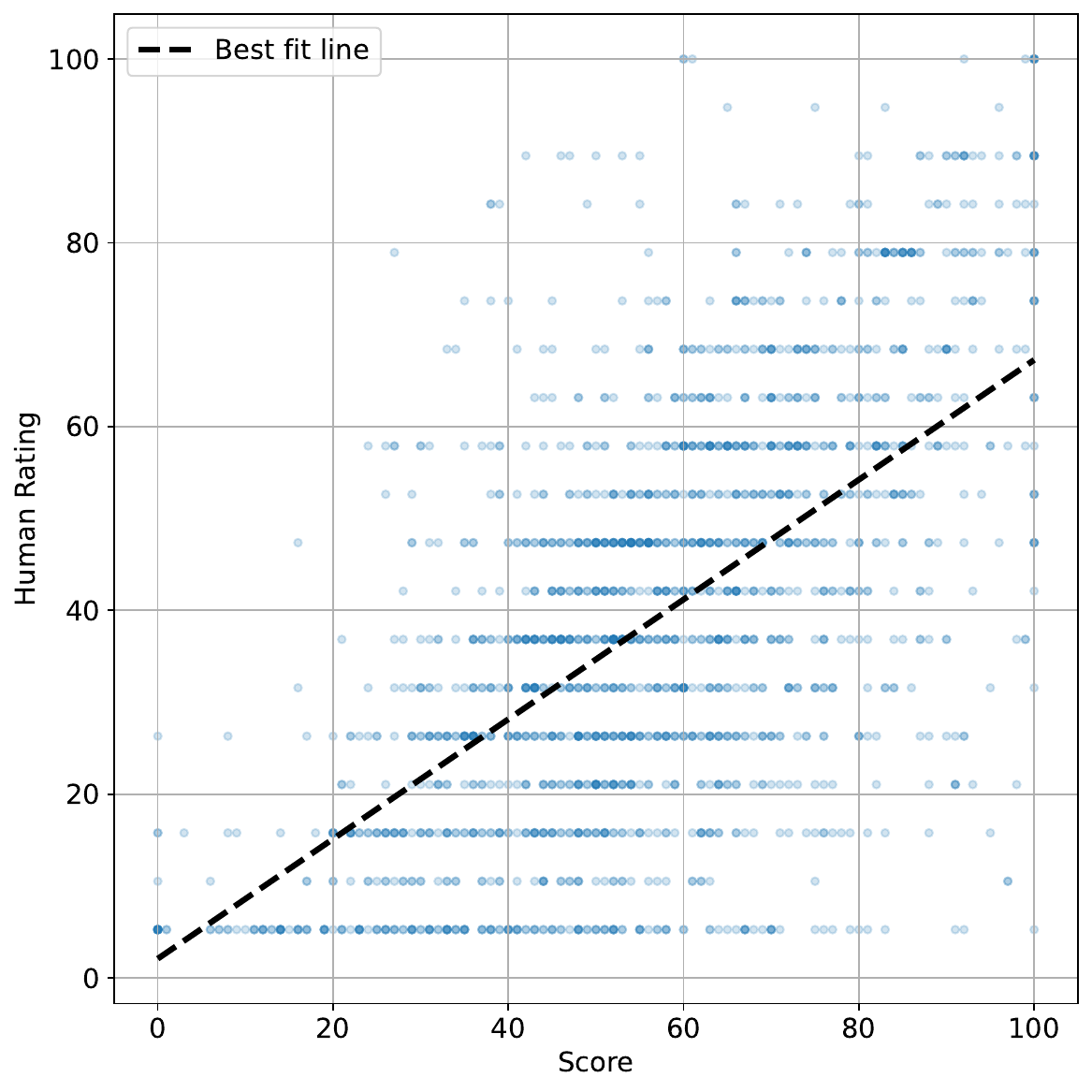}
    \caption{Scatter plot of scores and human ratings. Best fit line (linear regression) is shown in black.}
    \label{app:fig:human_rating_correlation}
\end{figure*}

Despite this, as is seen in Figure~\ref{app:fig:human_rating_correlation}, the CLIP-based score does not perfectly correlate with the human ratings. There are two reasons the Pearson correlation coefficient is not larger: (1) noise in the human rating responses and (2) deficiencies in the CLIP embeddings. While improving the image similarity metric is desirable, it was not the focus of our paper. For running ArtWhisperer, we only required a metric that worked reasonably well to generate a “reasonable score” (i.e., which we define as showing at least moderate correlation with recorded human rating assessments). Any subsequent analysis can of course use specialized similarity metrics on our data as we release all target and generated images. 

\begin{figure*}
    \centering
    \includegraphics[width=5.5in]{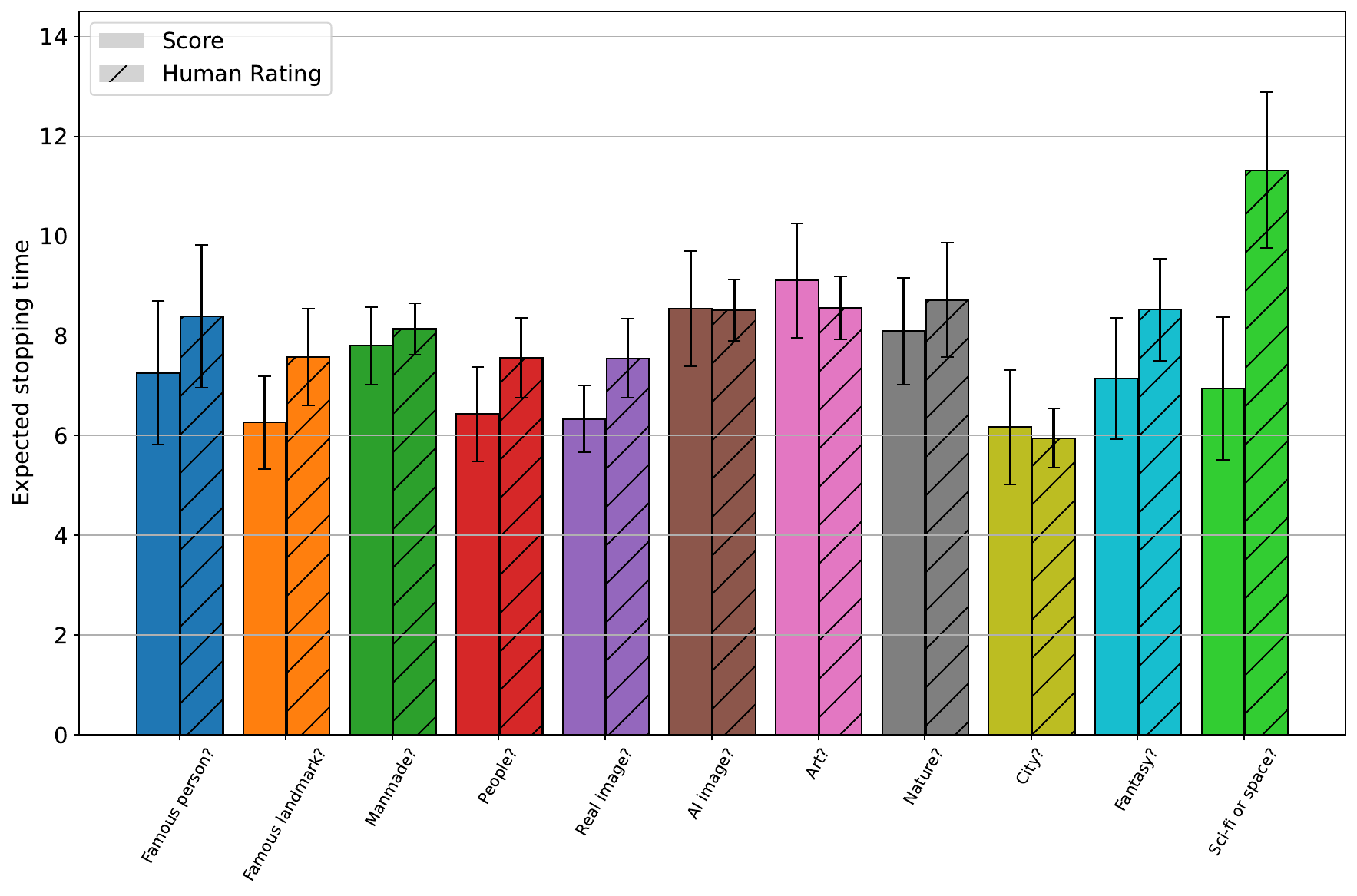}
    \caption{Comparing steerability across image groups when computing steerability using the score function and when using the human rating function.}
    \label{app:fig:human_rating_vs_score}
\end{figure*}

The results plotted in Figure~\ref{app:fig:human_rating_vs_score} also affirm CLIP as a reasonable choice for a metric – the score (based on the CLIP embedding) and human rating, while resulting in different steerability scores, offer similar conclusions across most image categories (e.g., the images that are AI-generated are less steerable than real images).

\subsection{Synthetic Prompters}\label{app:synthetic_prompter}
In this section, we describe a proof-of-concept Synthetic prompter based on the ArtWhisperer dataset. 
We fine-tuned a pretrained MT0-large model \cite{muennighoff2022crosslingual} using the IA3 method \cite{liu2022few}, training for 30 epochs and using a linearly decaying learning rate starting from $10^-3$ with the AdamW optimizer \cite{loshchilov2017decoupled}.

Our train dataset is based on the main ArtWhisperer dataset. We first randomly sampled 80\% of the ArtWhisperer dataset for training, with sampling taken over the target images (to ensure there are unseen images we can test on). For each user trajectory in the training set, we sampled a sliding window of user prompts. This is done to ensure the total prompt length of the model isn't too long to enable efficient training. Recalling that 50\% of the trajectories in the ArtWhisperer dataset have $\leq$ 5 total prompts, we select a window size of 6 as a reasonable cutoff (allowing a final prompt generation after seeing the 5 previous prompts).

Given that MT0-large is a text-only model, we need a way of encoding the image information for the model to simulate a user prompting on the given image. We opt to simply provide the model with the target prompt; after fine-tuning, the model uses the target prompt to condition its generation but does not repeat it verbatim. In addition to the target prompt, we also provide the model with a history of user prompts and scores indicating how well the image generated by a given prompt matches the target image. For training, we define a loss function on the next-prompt token probabilities to encourage the model to learn to predict subsequent prompts given the prompt and score histories. 

\begin{figure*}
    \centering
    \includegraphics[width=5.5in]{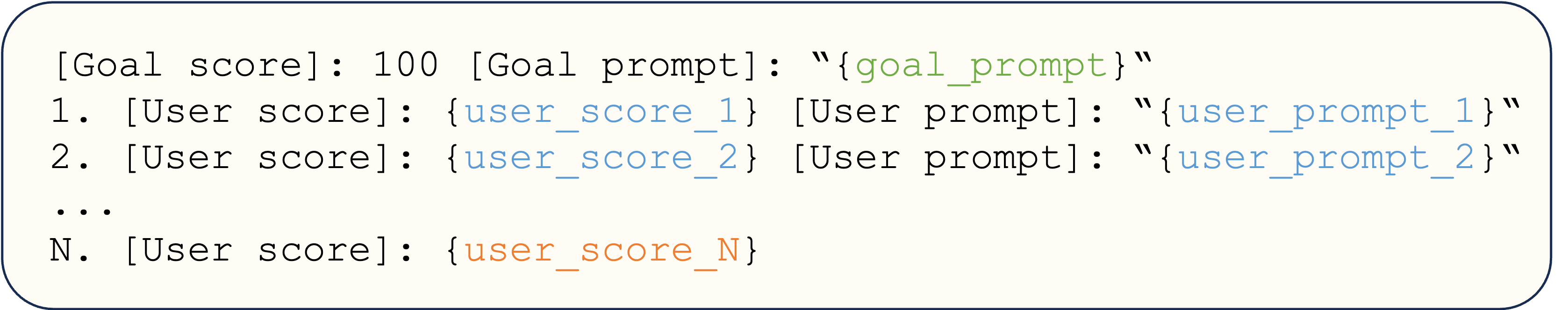}
    \caption{Prompt format for training and querying the synthetic prompter. $N \leq 6$ in our training and evaluation.}
    \label{app:fig:synthetic_prompt}
\end{figure*}

The prompt is shown in Figure~\ref{app:fig:synthetic_prompt}. The ``goal prompt'' (the target prompt) conditions the model on the target image content. The ``user\_score\_i'' values are set based on evaluating the distance between the target and generated image, though in general, we could use any metric of our choosing; the ``user\_prompt\_i'' values are set based on previously generated prompts. ``user\_score\_N'' is used to condition generation of a new prompt, and enables easy re-sampling by changing the ``score'' of the new prompt. Note this score is entered pre-evaluation, so we can set it to any value we like; in our evaluations, we set it to a randomly sampled value between 60 and 90.

This approach is similar to prior work like in Promptist \cite{hao2022optimizing}, where the authors fine-tune models to generate high-quality prompts given a human input through supervised learning and reinforcement learning approaches, or in \cite{zhu2023collaborative}, where the authors also use in-context learning approaches with demonstrations of the initial and final prompts in human trajectories. In both works, the goal is to optimize a prompt. Here, however, we seek to generate a trajectory of prompts that models how a human may behave. Thus while similar in principle, our objective function is different -- we train the model to generate the next prompt in a sequence rather than the best performing prompt from an initial prompt.

\subsubsection{Analysis of Synthetic Prompters}
\begin{figure*}
    \centering
    \includegraphics[width=5.5in]{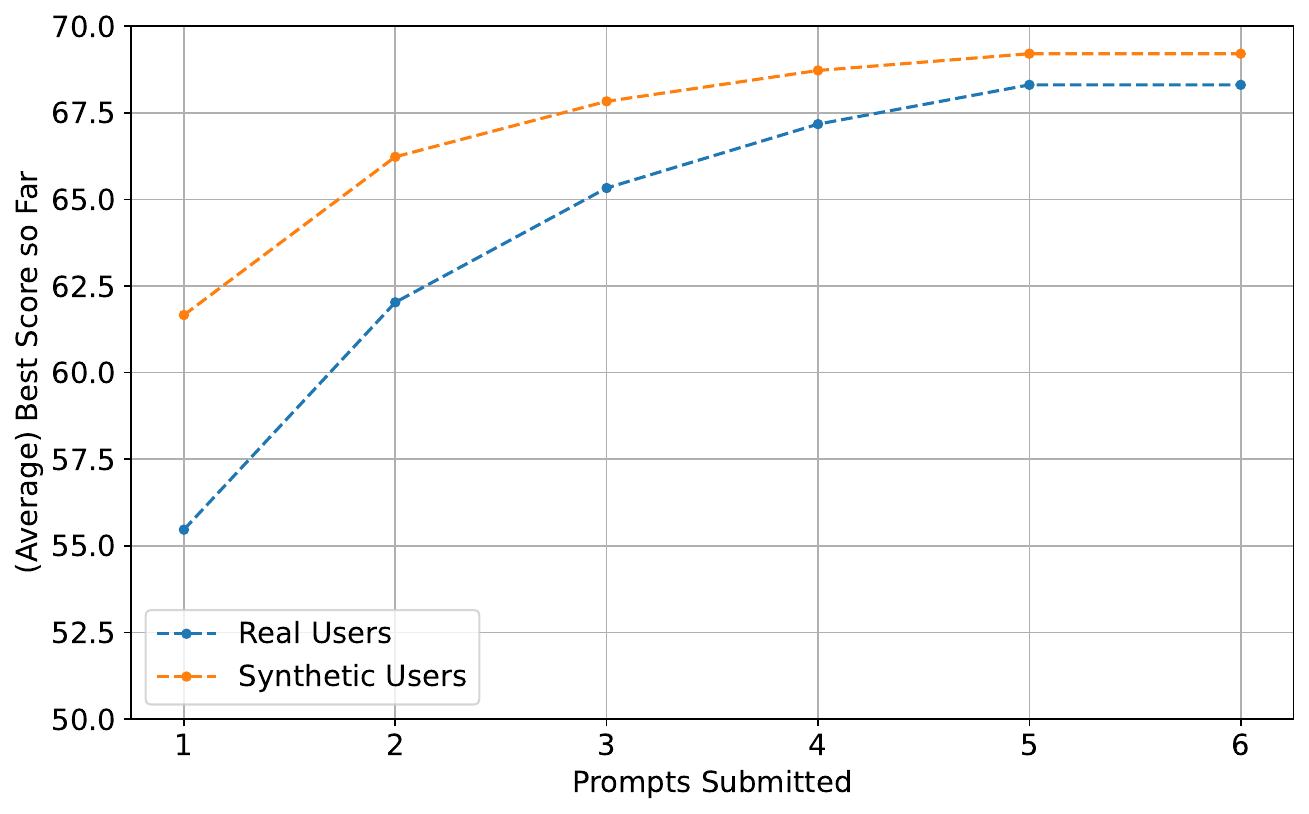}
    \caption{Comparing score progression for the synthetic prompter to real human users.}
    \label{app:fig:synthetic_prompter_compared_to_real}
\end{figure*}

Here we present an initial analysis that demonstrates the effectiveness of our dataset for training synthetic human prompters. After fine-tuning, we evaluate the synthetic prompter on the held-out test set. The target images in this test set (39 images) were excluded from all training data. For each image, we generate 10 sample trajectories. In Figure~\ref{app:fig:synthetic_prompter_compared_to_real}, we plot the average best score for the synthetic prompter and the real human prompters on the same test data. Note that while the synthetic prompter starts with a higher average, both approach a similar average score after 6 prompts. This suggests that the synthetic prompter indeed shares some similarities with the human prompters. 

\begin{figure*}
    \centering
    \includegraphics[width=5.5in]{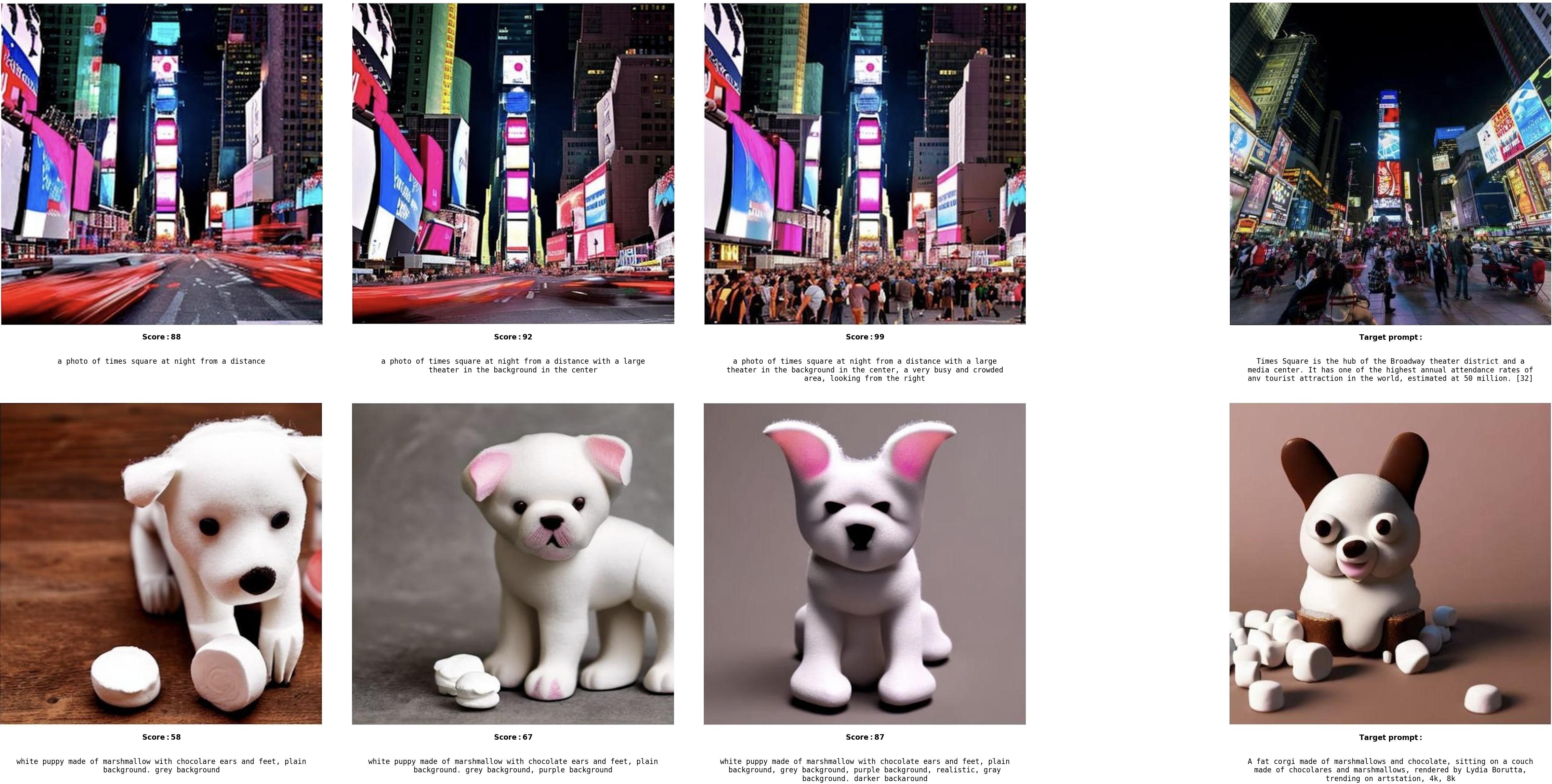}
    \caption{Example synthetic prompter trajectories. First 3 images show prompt progression for the synthetic prompter. The target image is shown in the rightmost column.}
    \label{app:fig:synthetic_prompter_trajectories}
\end{figure*}

In Figure~\ref{app:fig:synthetic_prompter_trajectories}, we plot two sample trajectories generated by the synthetic prompter. Note that across images, the prompter makes incremental changes; these changes are not restricted to appending phrases, but can involve substitutions and deletions throughout the prompt; this is in contrast to other automated prompters that exclusively complete text (these methods are not designed to simulate human behavior) \cite{hao2022optimizing}.

\subsection{Assessing VLMs for Feedback Utilization}\label{app:vlm_feedback_evaluation}
\begin{figure}
    \centering
    \includegraphics[width=5in]{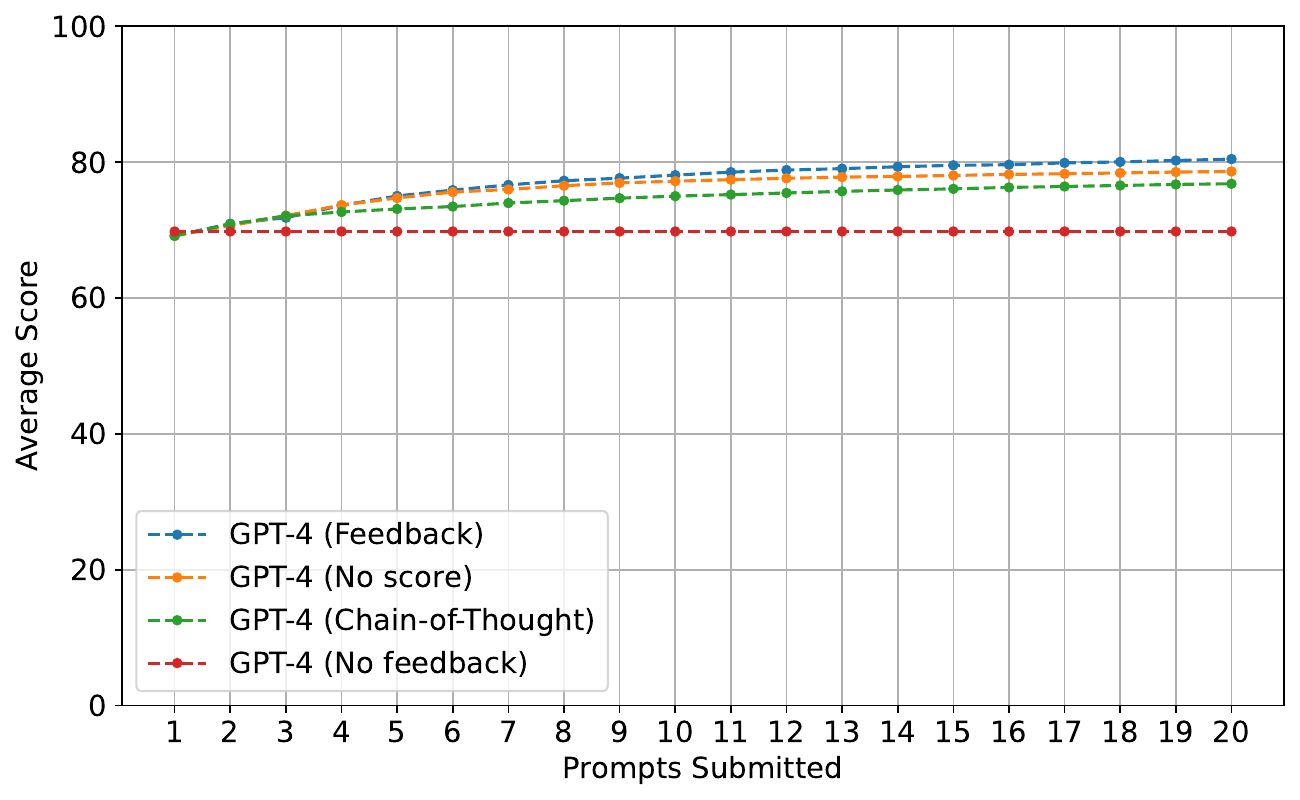}
    \caption{Comparison of additional prompting methods for GPT-4 playing the ArtWhisperer game.}
    \label{fig:gpt4_all_evaluation}
\end{figure}

In Figure~\ref{fig:gpt4_all_evaluation}, we evaluate several methods of prompting. This figure replicates the GPT-4 plots from Figure~\ref{fig:vlm_trajectory} and adds two additional prompting methods -- ``No score'' and ``Chain-of-thought.'' In ``No score,'' the model is provided feedback as normal, but only the generated image and not the ArtWhisperer score as well. In ``Chain-of-thought,'' the model is provided both the generated image and score feedback, but the VLM prompt used to update the old prompt used for SD uses a chain-of-thought reasoning approach. While all forms of feedback beat the baseline of no feedback, using a score and \textit{not} chain-of-thought reasoning has the highest performance. Interestingly, chain-of-thought reasoning, which typically increases model performance across a variety of tasks \cite{wei2022chain}, has lower performance here. Noting that performance for chain-of-thought only begins to diverge after 3 prompts have been submitted, we believe the decrease in performance may be due to the increased context length resulting from chain-of-thought reasoning. 

All the prompts used are included in Table~\ref{tbl:vlm_prompts}.

\begin{table}[ht]
\caption{Prompts used to have vision language models play ArtWhisperer.} 
\label{tbl:vlm_prompts}
\vskip 0.1in
\begin{center}
\begin{tabularx}{\textwidth}{|>{\hsize=0.7\hsize}X|>{\hsize=1.3\hsize}X|}
\hline
\multicolumn{2}{|c|}{\textbf{System Prompts}} \\ \hline
\textbf{Purpose} & \textbf{Prompt} \\ \hline
System prompt (Default) & You are playing a game using a text-to-image AI model. In the game, you will be shown a target image. Your goal is to write a prompt that, when input to the text-to-image model, generates a similar image to the target. You will be scored on the similarity between the generated image and the target image on a scale from 0-100. Your goal is to receive a 100/100.

You will first be shown the target image. The game will then proceed in rounds. Each round, you must write a prompt. You will then be shown the image generated by the text-to-image model using your prompt, as well as a score assessing its similarity to the target. You can use this feedback to update your prompt in the next round.

Referencing NSFW content will result in a black image and 0/100 score. Be descriptive yet concise. Each prompt must describe the target image in less than 50 words.
\\ \hline
System prompt (No score) & You are playing a game using a text-to-image AI model. In the game, you will be shown a target image. Your goal is to write a prompt that, when input to the text-to-image model, generates a similar image to the target.

You will first be shown the target image. The game will then proceed in rounds. Each round, you must write a prompt. You will then be shown the image generated by the text-to-image model using your prompt. You can use this feedback to update your prompt in the next round.

Referencing NSFW content will result in a black image. Be descriptive yet concise. Each prompt must describe the target image in less than 50 words.
\\ \hline
\hline
\multicolumn{2}{|c|}{\textbf{User Prompts}} \\ \hline
\textbf{Purpose} & \textbf{Prompt} \\ \hline
Initialization prompt (to get the first textual description of the target for all prompting approaches) & Target image: <image\_object>\\ \hline
Feedback and prompt updating (Feedback) & Using this prompt, the AI generated the following image: <image\_object> You received a score of <ArtWhisperer\_score>/100 for this image. Update your prompt to make the generated image closer to the target image.\\ \hline
Feedback and prompt updating (No score) & Using this prompt, the AI generated the following image: <image\_object> Update your prompt to make the generated image closer to the target image.\\ \hline
Feedback and prompt updating (Chain-of-Thought) & Using this prompt, the AI generated the following image: <image\_object> You received a score of <ArtWhisperer\_score>/100 for this image. Update your prompt to make the generated image closer to the target image. Reason step-by-step. First determine what is different between the generated and target images. Then update the prompt to better align the generated image with the target. Delimit the updated prompt with <prompt> tags.\\ \hline
\hline
\end{tabularx}
\end{center}
\vskip -0.1in
\end{table}

\end{document}